\documentclass[10pt,journal,compsoc]{IEEEtran}
\ifCLASSOPTIONcompsoc
  \usepackage[nocompress]{cite}
\else
  \usepackage{cite}
\fi
\ifCLASSINFOpdf
   \usepackage[pdftex]{graphicx}
\else
\fi
\usepackage{amsmath,amsfonts}
\usepackage{algorithmic}
\usepackage{algorithm}
\usepackage{array}
\usepackage[caption=false,font=normalsize,labelfont=sf,textfont=sf]{subfig}
\usepackage{textcomp}
\usepackage{stfloats}
\usepackage{url}
\usepackage{verbatim}
\usepackage{graphicx}
\usepackage{float}
\usepackage{subfig}
\usepackage{bm}
\usepackage{multirow}
\usepackage{makecell}
\usepackage{booktabs}
\usepackage[breaklinks]{hyperref}
\def\eg{\emph{e.g. }} 
\def\ie{\emph{i.e}. }

\usepackage[capitalize]{cleveref}
\crefname{section}{Sec.}{Secs.}
\Crefname{section}{Section}{Sections}
\Crefname{table}{Table}{Tables}
\crefname{table}{Tab.}{Tabs.}

\usepackage{xcolor}
\usepackage{colortbl}
\definecolor{color3}{RGB}{255, 255, 200}
\definecolor{color2}{RGB}{255, 220, 200}
\definecolor{color1}{RGB}{255, 181, 163}
\newcommand{\cc}[1]{\cellcolor{color#1}}

\begin{document}

\title{GUS-IR: Gaussian Splatting with Unified Shading for Inverse Rendering}

\author{
Zhihao Liang,~\IEEEmembership{Student~Member,~IEEE,}
Hongdong Li,
Kui Jia,
Kailing Guo and
Qi Zhang,
\IEEEcompsocitemizethanks{\IEEEcompsocthanksitem Zhihao Liang and Kailing Guo are with the School of Electronic and Information Engineering, South China University of Technology, Guangzhou 510641, China (e-mail: eezhihaoliang@mail.scut.edu.cn; guokl@scut.edu.cn).
\IEEEcompsocthanksitem Hongdong Li is with the Australian National University, Canberra 2600, Australia (e-mail: hongdong.li@anu.edu.au).
\IEEEcompsocthanksitem Kui Jia is with the School of Data Science, The Chinese University of Hong Kong, Shenzhen 518172, China (e-mail: kuijia@cuhk.edu.cn). 
\IEEEcompsocthanksitem Qi Zhang (corresponding author) is with VIVO, Xi'an 710055, China (e-mail: nwpuqzhang@gmail.com). \protect \\
}

\thanks{Manuscript received April 19, 2021; revised August 16, 2021.}}

\markboth{Journal of \LaTeX\ Class Files,~Vol.~14, No.~8, August~2021}%
{Shell \MakeLowercase{\textit{et al.}}: A Sample Article Using IEEEtran.cls for IEEE Journals}

\IEEEpubid{0000--0000/00\$00.00~\copyright~2021 IEEE}

\IEEEtitleabstractindextext{%
\begin{abstract}
Recovering the intrinsic physical attributes of a scene from images, generally termed as the inverse rendering problem, has been a central and challenging task in computer vision and computer graphics. In this paper, we present GUS-IR, a novel framework designed to address the inverse rendering problem for complicated scenes featuring rough and glossy surfaces.  This paper starts by analyzing and comparing two prominent shading techniques popularly used for inverse rendering, {\em forward shading} and {\em deferred shading}, effectiveness in handling complex materials.  More importantly, we propose a unified shading solution that combines the advantages of both techniques for better decomposition. In addition, we analyze the normal modeling in 3D Gaussian Splatting (3DGS) and utilize the shortest axis as normal for each particle in GUS-IR, along with a depth-related regularization, resulting in improved geometric representation and better shape reconstruction. Furthermore, we enhance the probe-based baking scheme proposed by GS-IR to achieve more accurate ambient occlusion modeling to better handle indirect illumination.
Extensive experiments have demonstrated the superior performance of GUS-IR in achieving precise intrinsic decomposition and geometric representation, supporting many downstream tasks (such as relighting, retouching) in computer vision, graphics, and extended reality.
\end{abstract}

}

\maketitle

\section{Introduction}
\label{sec:intro}
\IEEEPARstart{I}{nverse} rendering is a long-standing and challenging task in computer vision and computer graphics, aiming to recover intrinsic physical attributes (\eg scene geometry, surface material, and environment lighting) of a 3D scene from multiple observations. The main challenges arise from the optimization in uncontrolled environments with unknown illumination, which leads to ill-posedness in decomposing multiple strongly coupled physical attributes. Traditional methods~\cite{shen2013accurate, zheng2014patchmatch, galliani2015massively, schonberger2016structure, schonberger2016pixelwise, romanoni2019tapa, xu2020planar} struggled to simultaneously and accurately estimate the geometry, material, and illumination of a complex scene.

Recently, neural rendering techniques such as Neural Radiance Fields (NeRF) \cite{mildenhall2020nerf} have shown remarkable promise in scene geometric reconstruction tasks \cite{wang2021neus, yariv2021volume, wang2022neuris, yu2022monosdf, liang2023helixsurf}. However, most existing methods have overlooked the interaction between intrinsic physical attributes and illumination. Their implicit representations, which adopt an overly simplified appearance model as a function of view direction only, limit their applicability to many downstream tasks like relighting. Although recent methods \cite{zhang2021nerfactor, srinivasan2021nerv, jin2023tensoir, zhuang2024pre} introduce differentiable physical-based rendering to decompose physical attributes (\eg geometry, surface materials, and environmental lighting), they still face challenges related to low rendering speed and explicit editing, especially when it can not be rendered at interactive rates.

3D Gaussian Splatting (3DGS) \cite{kerbl20233d} has recently emerged as a promising technique to model 3D scenes and significantly boost the rendering speed to a real-time level. It combines explicit and more compact scene representation with differentiable rasterization \cite{zwicker2002ewa, zwicker2001surface}, resulting in a fast and remarkable performance for novel view synthesis. It is natural and essential to introduce physical-based rendering to 3DGS to achieve efficient inverse rendering tasks \cite{gao2023relightable, liang2024gs, jiang2024gaussianshader}. Despite their impressive relighting performance, these approaches still face challenges in geometric estimation and shading schemes. Firstly, during the 3DGS optimization, the adaptive control of the Gaussian density may lead to loose geometry, making it difficult to estimate accurate scene's normal. One key step in using 3DGS for inverse rendering is accurately representing the geometry to describe the transportation of illumination.
Although GS-IR~\cite{liang2024gs} introduces a depth-related regularization, it ignores the geometric attributes of the Gaussian ellipsoid by attaching a learnable vector as the normal for each particle, struggling to compactly associate the normal optimization with the geometry.
Consequently, it is necessary to model normal and better reconstruct geometry via the geometric attributes of the ellipsoid.

Besides, another critical aspect of employing 3DGS for inverse rendering involves the shading scheme.
Depending on where physical-based rendering is incorporated within the 3DGS-based framework, the shading scheme can be divided into forward shading \cite{jiang2024gaussianshader} (\ie shading each particle \textit{before} `splatting') and deferred shading \cite{liang2024gs} (\ie shading each pixel \textit{after} `splatting').
These different shading schemes have a substantial impact on material decomposition and overall rendering quality. Specifically, forward shading emphasizes the representation of diffuse colors, making it well-suited for complex scenes featuring rough surfaces, while deferred shading is more adept at capturing glossy objects with high specular regions. More details can be found in Fig. \ref{fig:forward_deferred_double}.
Thus, it is critical to define a suitable shading scheme to make the inverse rendering feasible for different cases, particularly for complex scenes with rough surfaces and specular regions.

\begin{figure*}
\centering
\includegraphics[width=1.0\linewidth]{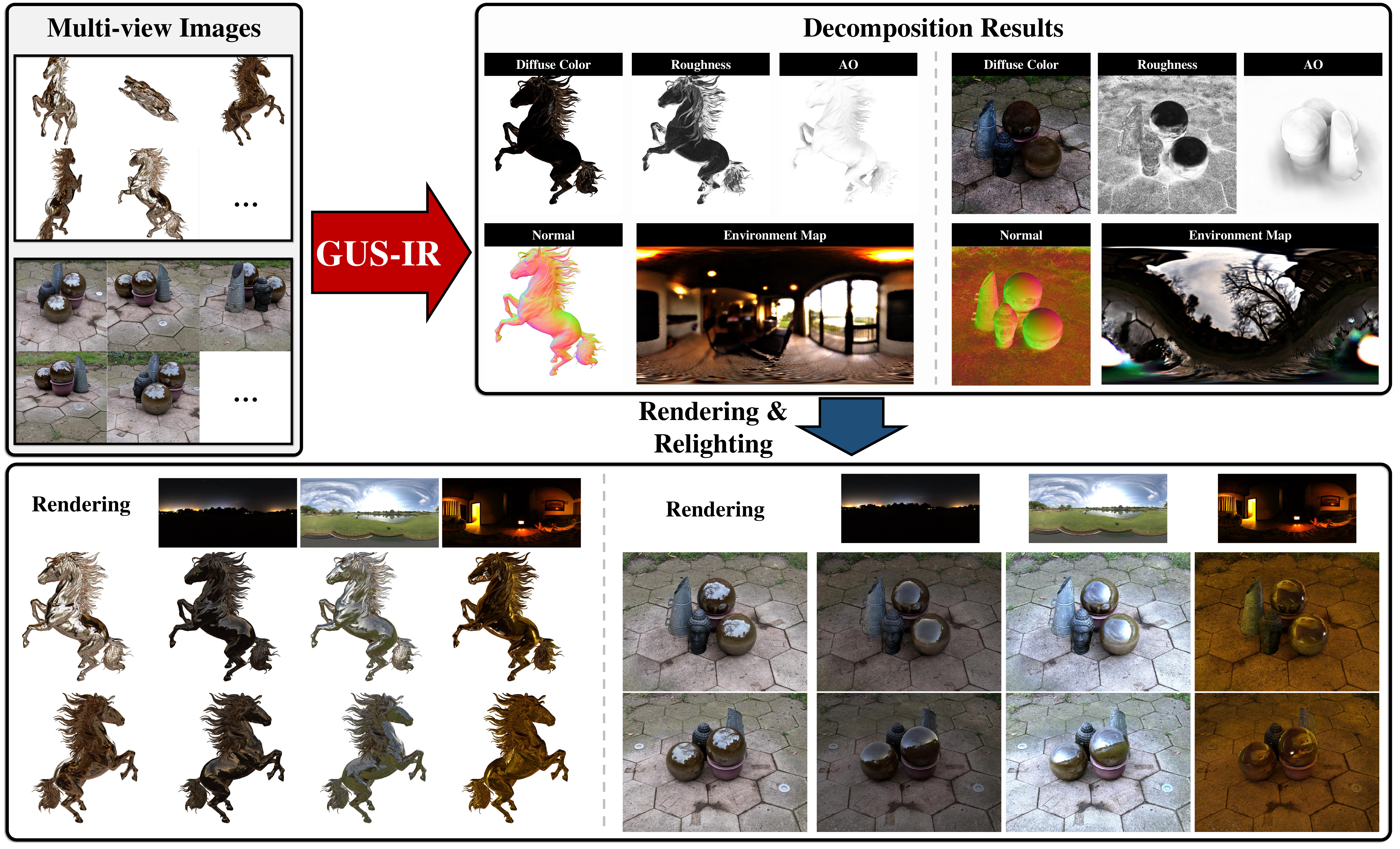}
\caption{Given multi-view captured images of a complex scene featuring rough and glossy surfaces, we propose \emph{GUS-IR} (\textbf{G}aussian Splatting with \textbf{U}nified \textbf{S}hading for \textbf{I}nverse \textbf{R}endering), which utilizes \textit{3D Gaussians} to recover high-quality physical properties (e.g., normal, material, illumination) under unknown illumination.
This enables us to perform advanced applications (\eg relighting), resulting in outstanding inverse rendering results.
Better viewed on screen with zoom-in, we successfully recovered the glossy surfaces of the marble balls placed in the center of the garden.
}
\label{fig:teaser}
\end{figure*}

In this paper, we are motivated to present a powerful 3DGS-based inverse rendering framework, GUS-IR (Gaussian Splatting with Unified Shading for Inverse Rendering), to achieve high-quality intrinsic decomposition under unknown illumination.
To represent the normal in 3DGS, we treat each particle as a surfel~\cite{huang20242d} and utilize its shortest axis as the orientation (\ie normal) for each particle.
Then we find that forward and deferred shading schemes are beneficial for reconstructing rough and glossy surface materials, respectively.
In GUS-IR, we thus unify both shading schemes to facilitate the intrinsic physical attributes decomposition of complex scenes containing various surface materials.
By leveraging the benefits of unification, GUS-IR addresses the challenges posed by glossy surfaces in complex scenes.
Additionally, we improve the baking scheme proposed in GS-IR, which leverages dense binary cubemaps to cache the occlusion during baking, achieving better ambient occlusion modeling.
The contributions of this work can be summarized as follows:
\begin{itemize}
\item We present \textit{GUS-IR} that unifies several shading mechanisms to achieve intrinsic decomposition of glossy objects and complex scenes;
\item We revisit forward and deferred shading schemes, and propose a unified shading scheme to handle complex scenes and better capture highlight details on glossy surfaces;
\item We propose the shortest-axis normal modeling to represent reliable geometry, along with the depth-related regularization for better reconstruction. 
\item We improve the probe-based baking solution proposed in GS-IR to better model ambient occlusion and handle indirect illumination.

\end{itemize}
We demonstrate the superiority of our method to baseline methods qualitatively and quantitatively on various challenging scenes, including the TensoIR Synthesis~\cite{jin2023tensoir}, Shiny Blender~\cite{verbin2022ref}, Glossy Blender~\cite{liu2023nero}, Mip-NeRF 360~\cite{barron2022mip}, and Ref-NeRF Real~\cite{verbin2022ref} datasets.

This work extends our preliminary work as described in ~\cite{liang2024gs}.
In this journal version, we first give a formal formulation of the rendering equation for Gaussians and analyze the advantages of two shading schemes (forward shading and deferred shading). 
We then propose a unified shading framework that systematically consolidates two shading mechanisms, which outperforms other shading methods on both diffuse and specular regions.
Finally, we replace the SH coefficients used in the baking solution of GS-IR with dense binary cubemaps to better handle ambient occlusion.
To validate this framework, we re-conducted all our experiments including new tests on glossy objects, achieving superior performance on view synthesis, relighting, and geometry reconstruction.

\section{Related Works}
\label{sec:formatting}
\subsection{Neural Representation and Rendering}
Recently, a surge of neural rendering techniques, exemplified by Neural Radiance Field (NeRF)~\cite{mildenhall2020nerf}, have succeeded in modeling light fields, supporting remarkable results in novel-view synthesis~\cite{barron2021mip, bi2020neural, fridovich2023k, fridovich2022plenoxels}, 3D and 4D reconstruction~\cite{liang2023helixsurf, wang2022neuris, yu2022monosdf, park2021nerfies}, generation~\cite{chan2021pi, poole2022dreamfusion, wang2024prolificdreamer}, and other advanced tasks~\cite{zhang2022arf, huang2022hdr, ma2022deblur}.
Instead of the triangular or quadrilateral meshes widely used in the traditional graphics pipeline, these methods adopt the implicit neural representation (INR) to model the geometry and appearance.
Specifically, for 3D reconstruction, NeuS~\cite{wang2021neus} and VolSDF \cite{yariv2021volume} use Multi-Layer Perceptron (MLP) to parameterize a signed distance field (SDF) to represent the geometry of the target object or scene, which can be friendly optimized as well as extracting an explicit mesh through iso-surface extraction algorithms~\cite{lorensen1987marching, ju2002dual}.
However, MLP incurs a substantial payload in rendering, which is further exacerbated by the ray-casting technique adopted by NeRF.
To address these inefficient, several methods proposed to use explicit structures to manage features and accelerate sampling, such as volumes~\cite{hedman2021baking, sun2022direct, fridovich2022plenoxels}, hash grids~\cite{muller2022instant}, tri-planes~\cite{chen2022tensorf, cao2023hexplane, fridovich2023k}, points~\cite{xu2022point} even mesh~\cite{huang2024sur2f}.
While these methods alleviate the payload, they struggle with high-fidelity real-time rendering, getting rid of inefficient ray-casting is essential to solve this dilemma.
To this end, NVDiffRec~\cite{shen2021deep, munkberg2022extracting, hasselgren2022shape} attempts to achieve efficient reconstruction using a mesh with a rasterization pipeline in the traditional form.
However, it struggles with complex topologies and scene reconstruction.
To address these challenges, 3DGS~\cite{kerbl20233d} incorporates explicit 3D Gaussian primitives with specially designed tile-based rasterization to achieve high-fidelity real-time rendering performance.
3DGS manages a set of particles, each of which maintains a 3D opacity field, and shades the pixel by parallelly `splatting' particles onto the screen space. 
Several contemporary works further fundamentally improve 3DGS~\cite{yu2024mip, liang2024analytic, huang20242d, yu2024gaussian, zhang2024rade} or apply it to advanced tasks~\cite{wu20244d, gao2024gaussianflow}.
For inverse rendering, GS-IR~\cite{liang2024gs}, GaussianShader~\cite{jiang2024gaussianshader}, and RelightableGS~\cite{gao2023relightable} achieve promising results based on 3DGS.
For modeling the normal in 3DGS, GS-IR and RelightableGS attach a learnable orientation for each particle as the normal, and GaussianShader associates a predicted residual on the shortest axis as the normal.
Different from them, we follow 2DGS~\cite{huang20242d}, treating each particle as a flat surfel and using the shortest axis as the normal in GUS-IR.
Our experimental results validate these normal representations and show that our scheme is superior to other schemes in terms of both decomposition and relighting results.

\subsection{Inverse Rendering}
Inverse rendering aims to decompose the physical attributes (\eg geometry, materials, and illumination conditions) from observations.
Importantly, in inverse rendering tasks, appearance tends to be treated as the physical interaction~\cite{cook1982reflectance, kajiya1986rendering, walter2007microfacet} of the surface and illumination, rather than a fitted spatially view-dependent function.
However, the inherent ambiguity between observation and the underlying properties makes this problem extremely ill-posed.
To alleviate the challenge, some approaches propose constraints to obtain well-defined illumination conditions, such as fixing illumination conditions, rotating the target object, or co-locating the light source with the moving camera~\cite{dong2014appearance, xia2016recovering, bi2020deep, bi2020deep2, luan2021unified, nam2018practical}.
Inspired by the success of NeRF, recent approaches~\cite{bi2020neural, srinivasan2021nerv, zhang2021physg, boss2021nerd, boss2021neural, zhang2021nerfactor, zhang2022modeling, hasselgren2022shape, zhang2022iron, jin2023tensoir} leverage neural representation and rendering to alleviate the ill-posedness in the optimization and aim to simulate physical interactions from multiple observations under unknown conditions.
As pioneers, NRF~\cite{bi2020neural} and NeRD~\cite{boss2021nerd} follow NeRF and use an MLP to parameterize the physical field (\eg volumetric density, normal, BRDF) of the target scene.
Instead of a view-dependent MLP, they conduct the physical-based rendering equation for shading.
NeRV~\cite{srinivasan2021nerv}, NeRFactor~\cite{zhang2021nerfactor}, InvRender~\cite{zhang2022modeling} further model the visibility or shadows in the target scene.
In addition, PhySG~\cite{zhang2021physg} uses Spherical Gaussians (SGs) to represent illumination and BRDFs, TensoIR~\cite{jin2023tensoir} adopts the tensor factorization~\cite{chen2022tensorf} to compute visibility and indirect illumination efficiently.
However, these methods are limited to the object-level and almost require several hours even days for inverse rendering.
Based on 3DGS, GaussianShader~\cite{jiang2024gaussianshader} handle the decomposition of reflective objects, GS-IR~\cite{liang2024gs} and RelightableGS~\cite{gao2023relightable} respectively leverage precomputation and ray-tracing techniques to handle the visibility.
Notably, the above methods can be easily transferred to complex scenes for inverse rendering.
In this work, we propose a 3DGS-based framework to decompose the geometry, materials, and illumination conditions for rough and glossy objects, even complex scenes.

\section{3DGS for Inverse Rendering}
\label{sec:preliminary}
In this section, we review GS-IR \cite{liang2024gs} and give the technical backgrounds and math symbols that are necessary for the presentation of our proposed method.

\subsection{3D Gaussian Splatting}
3D Gaussian Splatting (3DGS) \cite{kerbl20233d} explicitly represents a 3D scene as a set of particles, each particle stores the parameters that express a 3D Gaussian signal ${g}$ with a view-dependent color $\bm{c}$ modeled by spherical harmonics (SH), a mean (\ie position) $\bm{\mu} \in \mathbb{R}^3$ and an anisotropic covariance matrix $\bm{\Sigma} \in \mathbb{R}^{3 \times 3}$, where the covariance matrix $\bm{\Sigma} = \bm{R} \bm{S} \bm{S}^\top \bm{R}^\top$ is obtained by a rotation matrix $\bm{R}$ and scaling $\bm{S}$.
$\bm{S}$ and $\bm{R}$ refer to a diagonal matrix $\mathrm{diag}(s_x, s_y, s_z)$ and a rotation matrix constructed from a unit quaternion $\bm{q}$.
During rasterization, 3DGS conducts perspective projection and splats the 3D Gaussian signal $g$ into a 2D Gaussian signal $\tilde{g}$ with mean vector $\bm{\tilde{\bm{\mu}}}$ and covariance $\tilde{\bm{\Sigma}}$ on the screen space according to the extrinsic matrix $\bm{T}$ and intrinsic matrix $\bm{K}$.
After obtaining all 2D Gaussian signals in screen space, 3DGS shades the pixel via $\alpha$-blending $N$ 2D Gaussian signals covering the pixel $\bm{u}$, and we finally obtain the color $\hat{\bm{c}}(\bm{u})$:
\begin{equation}
\begin{aligned}
\alpha_k &= \tilde{g}(\bm{u} | \tilde{\bm{\mu}}_k, \tilde{\bm{\Sigma}}_k)o_k,\quad
T_k = \prod^{k - 1}_{j = 1}(1 - \alpha_j), \\
\hat{\bm{c}}(\bm{u}) &= \sum^N_{k=1} w_k \bm{c}_k,\quad
w_k = T_k \alpha_k \\
\end{aligned}
\label{eq:volrend}
\end{equation}
where $o_k$ and $\bm{c}_k$ are the opacity and color associated with the $k$-th point.

\subsection{Gaussian Splatting for Inverse Rendering}
For inverse rendering, some methods~\cite{liang2024gs, jiang2024gaussianshader, gao2023relightable} represent the view-dependent appearance as the physical interaction between illumination and materials, rather than spherical harmonics used in 3DGS.
To model the appearance of a given surface point $\bm{x}$ with normal $\bm{n}$ from view direction $\bm{v}$, they follow the classic rendering equation~\cite{kajiya1986rendering} to formulate the outgoing radiance $L_o$ via integrating the response of the incident radiance over the upper hemisphere $\Omega$:
\begin{equation}
\begin{aligned}
L_o(\bm{x}, \bm{v}) &= \int_\Omega L_i(\bm{x}, \bm{l}) f_r(\bm{x}, \bm{l}, \bm{v}) (\bm{l} \cdot \bm{n}) d\bm{l}, \\
f_r(\bm{x}, \bm{l}, \bm{v}) &= f_r^\text{d}(\bm{x}) + f_r^\text{s}(\bm{x}, \bm{l}, \bm{v}),
\end{aligned}
\label{eq:rendering}
\end{equation}
where $f_r$ describes the Bidirectional Reflectance Distribution Function (BRDF), mostly formulated by the Cook-Torrance microfacet model~\cite{cook1982reflectance, walter2007microfacet}.
$f_r$ can be divided into diffuse term $f_r^\text{d}$ and specular term $f_r^\text{s}$ depending on whether related to incidence $\bm{l}$ and view direction $\bm{v}$.
In different workflows, the reflection in $f_r$ is modeled by different physical attributes, such as Dielectric F0 coefficients with metallic~\cite{liang2024gs}, and specular color~\cite{jiang2024gaussianshader}.
In GUS-IR, for each particle, we define learnable diffuse color $\bm{\alpha}$ and specular color $\bm{s}$ to represent its base color and reflection color, respectively.
After dividing the $f_r$ into diffuse and specular terms, the rendering equation \cref{eq:rendering} can be rewritten as the combination of diffuse component $L_o^\text{d}$ and specular component $L_o^\text{s}$ as:
\begin{equation}
\label{eq:rendering_rewrite}
\begin{aligned}
L_o(\bm{x}, \bm{v}) =& L_o^\text{d}(\bm{x}) + L_o^\text{s}(\bm{x}, \bm{v}) \\
L_o^\text{d}(\bm{x}) =& f_r^\text{d}(\bm{x}) \int_\Omega L_i(\bm{x}, \bm{l}) (\bm{l} \cdot \bm{n}) d\bm{l}, \\
L_o^\text{s}(\bm{x}, \bm{v}) =& \int_\Omega L_i(\bm{x}, \bm{l}) f^\text{s}_r(\bm{x}, \bm{l}, \bm{v}) (\bm{l} \cdot \bm{n}) d\bm{l}.
\end{aligned}
\end{equation}

Specifically, previous methods almost adopt the image-based lighting (IBL) model and assume that direct illumination comes from a distance, thus incident radiance $L_i(\bm{x}, \bm{l})$ degenerates as $L_i(\bm{l})$ that is only related to incidence $\bm{l}$, and the direct light sources are represented by an HDRI.
They further use split-sum approximation~\cite{karis2013real} to solve the intractable integral in \cref{eq:rendering_rewrite} as:
\begin{equation}
\label{eq:ibl_splitsum}
\begin{aligned}
L_o^\text{d}(\bm{x}) =& f_r^\text{d}(\bm{x}) \underbrace{
\int_\Omega L_i(\bm{l}) (\bm{l} \cdot \bm{n}) d\bm{l}
}_{I^\text{d}}, \\
L_o^\text{s}(\bm{x}, \bm{v})
=& \int_\Omega L_i(\bm{l}) f^\text{s}_r(\bm{x}, \bm{l}, \bm{v}) (\bm{l} \cdot \bm{n}) d\bm{l} \\
\approx&
\underbrace{
\int_\Omega f^\text{s}_r(\bm{x}, \bm{l}, \bm{v}) (\bm{l} \cdot \bm{n}) d\bm{l}
}_{R}
\underbrace{
\int_\Omega p(\bm{l}, \bm{v}) L_i(\bm{l}) (\bm{l} \cdot \bm{n}) d\bm{l}
}_{I^\text{s}},
\end{aligned}
\end{equation}
where $p(\bm{l}, \bm{v})$ is related to the roughness of surface point $\bm{x}$ and denotes the importance (\ie probability density function, PDF) of incidence $\bm{l}$ over the upper hemisphere $\Omega$.
Note that $I^\text{d}$ and $I^\text{s}$ denote the incident radiance integral over the upper hemisphere and the reflected lobe.
$I^\text{d}$, $I^\text{s}$, and $R$ can be precomputed in advance from the HDRI and stored in look-up tables.

\begin{figure*}[!t]
\centering
\includegraphics[width=1.\textwidth]{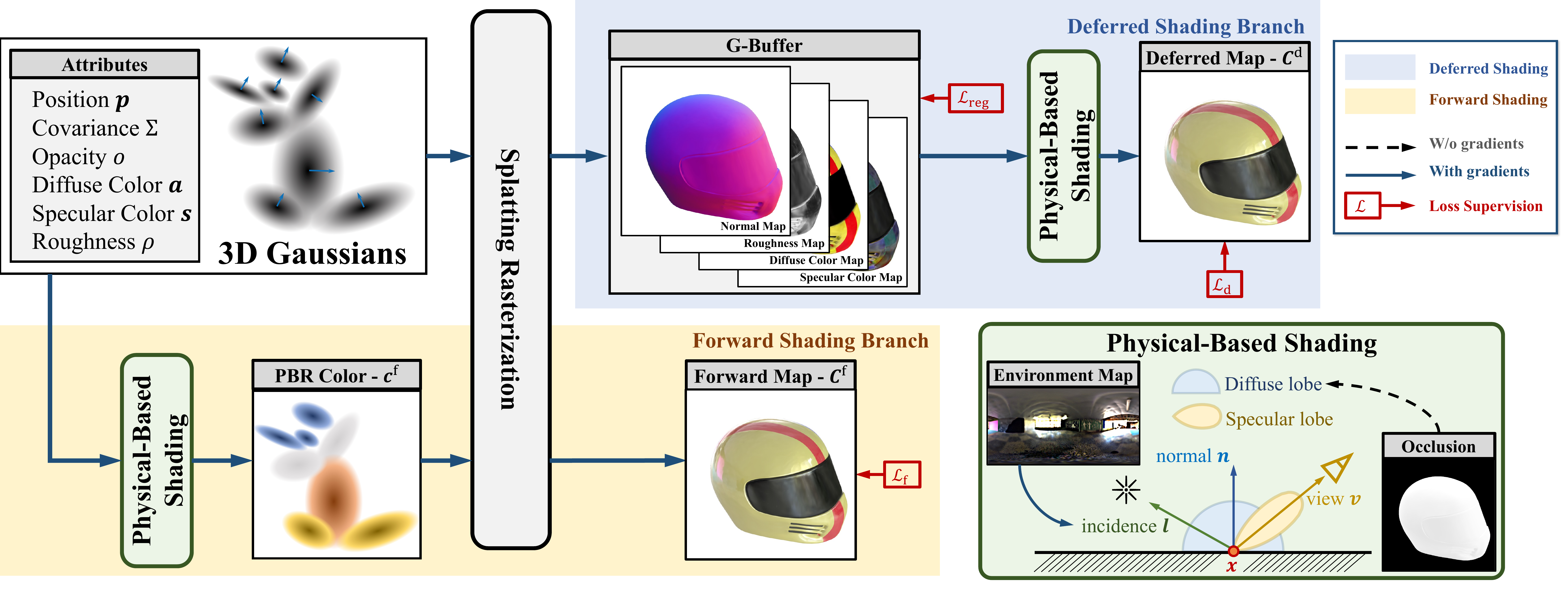}
\caption{Overview of GUS-IR.
During optimization, GUS-IR simultaneously conducts forward and deferred shading schemes and supervises the rendering results produced by both schemes.
We use the shortest axis towards the view as the particle's normal for forward shading and render a normal map for deferred shading.
}
\label{fig:pipeline}
\end{figure*}

\begin{figure*}[t]
\centering
\includegraphics[width=0.95\linewidth]{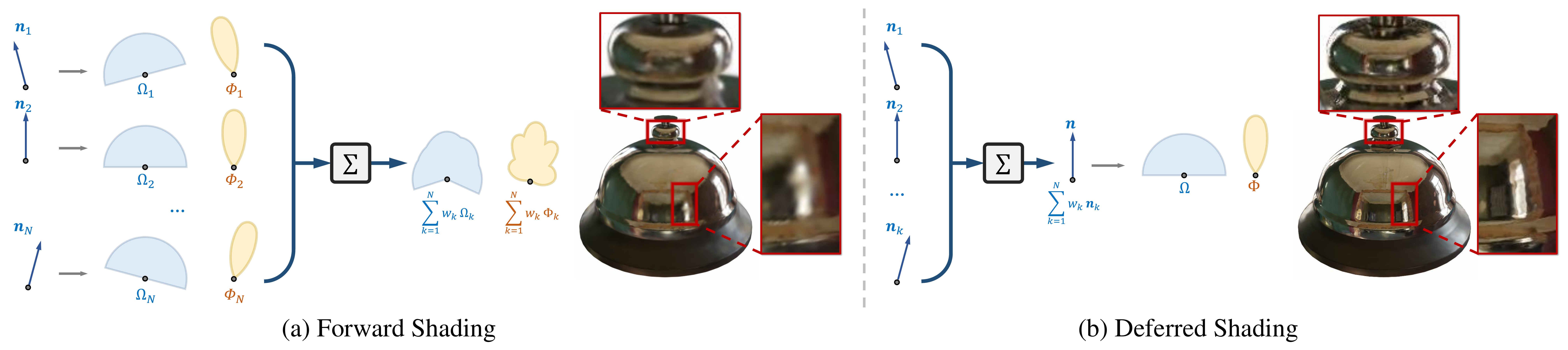}
\caption{Difference between forward and deferred shading schemes. $\Sigma$ denotes the accumulation in the volume rendering. $\bm{n}_k, \Omega_k, \Phi_k$ are the normal, upper hemisphere, and BRDF lobe of the $k$-th particle, respectively.
The \textit{table bell} examples show that \textbf{deferred shading} can better capture and reconstruct glossy details than \textbf{forward shading}.
}
\label{fig:forward_deferred_double}
\end{figure*}

In addition, GS-IR treats the irradiance $I^\text{d}$ in the diffuse component as the combination of direct irradiance $I^\text{\tiny d-dir}$ and indirect irradiance $I^\text{\tiny d-indir}$:
\begin{equation}
\begin{aligned}
I^\text{d}(\bm{x}) =& \int_\Omega L_i(\bm{x}, \bm{l}) (\bm{l} \cdot \bm{n}) d\bm{l} \\
=& \int_{\Omega_\text{V}} L^\text{\tiny dir}_i(\bm{x}, \bm{l}) (\bm{l} \cdot \bm{n}) d\bm{l} +
\int_{\Omega_\text{O}} L^\text{\tiny indir}_i(\bm{x}, \bm{l}) (\bm{l} \cdot \bm{n}) d\bm{l} \\
\approx& \left(1 - \text{O}(\bm{x})\right) I^\text{\tiny d-dir}(\bm{x}) + \text{O}(\bm{x}) I^\text{\tiny d-indir}(\bm{x}),
\label{eq:diffuse}
\end{aligned}
\end{equation}
where $\Omega_\text{V}$ and $\Omega_\text{O}$ refer to the visible and occluded domain on the upper hemisphere $\Omega$, respectively.
In GS-IR, the ambient occlusion $\text{O}(\bm{x})$ is pre-calculated and cached in occlusion volumes $\mathcal{V}^\text{O}$ during the baking stage, which can be efficiently recovered in the decomposition stage.
In the baking stage, GS-IR freezes the geometry and regularly places occlusion volumes $\mathcal{V}^\text{O}$ in the bounded 3D space.
Each volume $\bm{v}^\text{O} \subset \mathcal{V}^\text{O}$ first captures a depth cubemap $\{\bm{D}^{\bm{v}^\text{O}}_i\}^6_{i=1}$ and then converts it into a binary occlusion cubemap $\{\bm{O}^{\bm{v}^\text{O}}_i\}^6_{i=1}$ through a manually set distance threshold, finally cache the low-frequent occlusion in the form of spherical harmonics (SH) coefficients $\bm{f}^\text{O}_i$.
In summary, the ambient occlusion of point $\bm{x}$ can be expressed as:
\begin{equation}
O(\bm{x}) = \sum^\mathit{deg}_{l=0} \sum^{l}_{m=-l} \bm{f}^\text{O}_{\bm{x}(lm)} Y_{lm}(\bm{n}),
\label{eq:occlusion}
\end{equation}
where $\bm{n}$ refers to the normal of point $\bm{x}$, $\mathit{deg}$ denotes the degree of SH, and $\{Y_{lm}(\cdot)\}$ is a set of real basis of SH. Notably, $\bm{f}^\text{O}_{\bm{x}}$ are the SH coefficients at point $\bm{x}$, which is obtained by conducting normal-aware masked-trilinear interpolation in the decomposition stage.

\section{Method}
\label{sec:method}
In this section, we present our novel framework GUS-IR for the inverse rendering of glossy objects and complex scenes, as shown in \cref{fig:pipeline}. Given a set of images with unknown illumination, our GUS-IR recovers Gaussians with Unified Shading for Inverse Rendering, achieving impressive intrinsic decomposition and relighting on both glossy objects and complex scenes.
Specifically, we first discuss different ways of normal modeling in \cref{subsec:normal}.
Next, we revisit the rendering equation and propose the unification of forward and deferred shading schemes in \cref{subsec:unification}.
Then, we introduce an improved baking strategy that models ambient occlusion for handling indirect illumination more accurately compared to GS-IR in \cref{subsec:occlusion}.
Finally, we summarize the supervision of the GUS-IR in \cref{subsec:losses}.

\subsection{Normal Modeling}
\label{subsec:normal}
The first key problem faced by introducing 3DGS for inverse rendering is reconstructing geometry, and normal estimation is crucial.
Since each particle in 3DGS models an outward attenuating opacity field instead of a well-defined geometry, 3DGS fails to natively represent normal.
To overcome this limitation, previous methods~\cite{liang2024gs, gao2023relightable} attach each particle with a learnable normal.
In contrast to them, GaussianShader~\cite{jiang2024gaussianshader} proposes to add a learnable residual on the shortest axis of the particle as the normal.

In GUS-IR, we directly treat the shortest axis of the particle as normal to achieve a tight geometric relation.
To supervise the learning of normal in GUS-IR, we first adopt the linear interpolation scheme proposed in GS-IR to produce a reliable depth map and derive a pseudo-normal map for supervision.
Then, unlike GS-IR, we directly run differentiable physical-based rendering in the first stage to better handle glossy surfaces.
Please note that we do not stop the gradients of normal and the depth map in these supervisions like GS-IR.
Experiments in \cref{subsec:ablation} validate these normal representations and demonstrate the effectiveness of our adopted scheme for geometry reconstruction.

\subsection{Shading Unification for Inverse Rendering}
\label{subsec:unification}
In inverse rendering, the view-dependent appearance is modeled as the physical interaction between illumination and materials as mentioned at \cref{sec:preliminary}.
It is straightforward to use the physical-based rendering function (\cref{eq:rendering}) to shade each particle, then perform volume rendering to accumulate the color of the pixel, which is called \emph{forward shading}.
According to \cref{eq:rendering} and \cref{eq:volrend}, given $N$ particles splatted on the pixel $\bm{u}$, the result of forward shading $\hat{\bm{c}}_\text{for}(\bm{u})$ can be formulated as:
\begin{equation}
\label{eq:forward_shading}
\hat{\bm{c}}_\text{for}(\bm{u}) =
\sum^N_{k=1} w_k
\int_{\Omega_k} L_i(\bm{\mu}_k, \bm{l}) f_r(\bm{\mu}_k, \bm{l}, \bm{v}) (\bm{l} \cdot \bm{n}_k) d\bm{l},
\end{equation}
where $\Omega_k$ denotes the upper hemisphere of the $k$-th particle with position $\bm{\mu}_k$ and normal $\bm{n}_k$.
In contrast, \emph{deferred shading} first leverages volume rendering to obtain the geometric and physical attributes of pixel $\bm{u}$, then obtain the pixel color $\hat{\bm{c}}_\text{for}(\bm{u})$ through a one-times shading:
\begin{equation}
\label{eq:deferred_shading}
\begin{aligned}
\bm{n} =& \sum^N_{i=k}w_k \bm{n}_k,\quad
\bm{x} = \sum^N_{i=k}w_k \bm{\mu}_k,\quad
\\
\hat{\bm{c}}_\text{def}(\bm{u}) =&
\int_\Omega 
L_i(\bm{x}, \bm{l})
\left(
\sum^N_{i=k}w_k f_r(\bm{\mu}_k, \bm{l}, \bm{v})
\right)  (\bm{l} \cdot \bm{n}) d\bm{l},
\end{aligned}
\end{equation}
where $\bm{x}$ and $\bm{n}$ are the expected position and normal of the pixel $\bm{u}$, the upper hemisphere $\Omega$ is centered at $\bm{x}$ around $\bm{n}$.

\begin{figure*}[t]
\centering
\includegraphics[width=\linewidth]{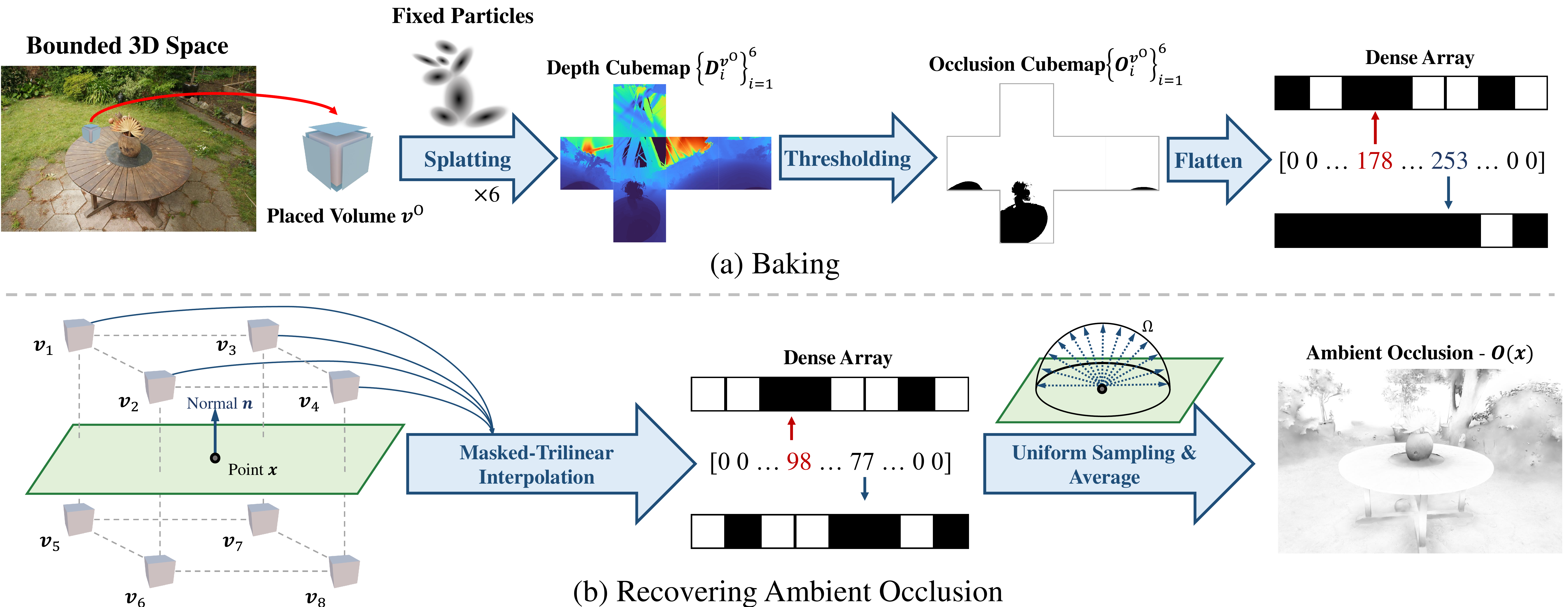}
\caption{Overview of caching occlusion in the baking stage and recovering occlusion in the decomposition from volumes.
We employ the dense structure to bake occlusion volumes for modeling ambient occlusion.
For each volume $\bm{v}^\text{O}$, we run six passes to render the depth cubemap $\{\bm{D}^{\bm{v}^\text{O}}_i\}^6_{i=1}$ and convert the cubemap into a binary occlusion cubemap $\{\bm{O}^{\bm{v}^\text{O}}_i\}^6_{i=1}$ through a manually set distance threshold.
Then we flatten the binary occlusion cubemap and store it in a dense byte array.
Given a surface point $\bm{x}$ with normal $\bm{n}$, we conduct normal-aware masked-trilinear interpolation to get the occlusion cubemap, averaging the uniform sampling occlusion as the ambient occlusion $O(\bm{x})$.
}
\label{fig:baking}
\end{figure*}

To better deduce the difference in shading results, we only consider direct illumination and adopt the IBL model to simplify the lighting condition.
Then we assume that $\sum^N_{k=1} w_k = 1$ in volume rendering and all particles have the same material (\ie spatially constant).
Given these conditions, the indirect radiance $L_i(\bm{x}, \bm{l})$ degenerates as $L_i(\bm{l})$ that is only related to incidence $\bm{l}$, and the BRDFs of all particles can be express as $f_r(\bm{l}, \bm{v})$.
In this way, the shading results in \cref{eq:forward_shading} and \cref{eq:deferred_shading} can be rewritten as:
\begin{equation}
\label{eq:shading_results1}
\begin{aligned}
\hat{\bm{c}}_\text{for}(\bm{u}) =&
\sum^N_{k=1} w_k
\int_{\Omega_k} L_i(\bm{l}) f_r(\bm{l}, \bm{v}) (\bm{l} \cdot \bm{n}_k) d\bm{l}, \\
\hat{\bm{c}}_\text{def}(\bm{u}) =&
\int_\Omega 
L_i(\bm{l}) \left(
\sum^N_{i=k} w_k f_r(\bm{l}, \bm{v})
\right)
(\bm{l} \cdot \bm{n}) d\bm{l} \\
=& \int_\Omega 
L_i(\bm{l}) f_r(\bm{l}, \bm{v}) (\bm{l} \cdot \bm{n}) d\bm{l}.
\end{aligned}
\end{equation}

Obviously, the result of deferred shading converges to \cref{eq:rendering}, while forward shading still maintains the form of volume rendering in \cref{eq:volrend}.
We further decompose them into diffuse and specular terms: $f_r(\bm{l}, \bm{v}) = f^\text{d}_r + f^\text{s}_r(\bm{l}, \bm{v})$.
Note that the $f^\text{d}_r$ is constant under our assumption.
The forward shading result $\hat{\bm{c}}_\text{for}(\bm{u})$ can be rewritten as:
\begin{equation}
\label{eq:forward_approx}
\begin{aligned}
\hat{\bm{c}}_\text{for}(\bm{u}) =&
\sum^N_{k=1} w_k
\int_{\Omega_k} L_i(\bm{l}) (f^\text{d}_r + f^\text{s}_r(\bm{l}, \bm{v})) (\bm{l} \cdot \bm{n}_k) d\bm{l} \\
=& \sum^N_{k=1} w_k f^\text{d}_r \int_\Omega L_i(\bm{l}) (\bm{l} \cdot \bm{n}_k) d\bm{l} + \\
& \sum^N_{k=1} w_k \int_\Omega L_i(\bm{l}) f^\text{s}_r(\bm{l}, \bm{v}) (\bm{l} \cdot \bm{n}_k) d\bm{l} \\
\approx& f^\text{d}_r \sum^N_{k=1} w_k I^\text{d}_\text{k} + \sum^N_{k=1}w_k R_k I^\text{s}_k, \\
=& \hat{\bm{c}}^\text{d}_\text{for}(\bm{u}) + 
\hat{\bm{c}}^\text{s}_\text{for}(\bm{u}),
\end{aligned}
\end{equation}
where $I^\text{d}_k$, $I^\text{s}_k$ and $R_k$ respectively denote the relevant precomputed results in the split-sum approximation of the $k$-th particle.
The deferred shading result in~\cref{eq:shading_results1} is convergence to surface shading, thus we directly adopt the split-sum approximation (\cref{eq:ibl_splitsum}) to decompose the results as:
\begin{equation}
\label{eq:deferred_approx}
\begin{aligned}
\hat{\bm{c}}_\text{def}(\bm{u}) =& \int_\Omega 
L_i(\bm{l}) (f^\text{d}_r + f^\text{s}_r(\bm{l}, \bm{v})) (\bm{l} \cdot \bm{n}) d\bm{l}, \\
=& f^\text{d}_r \int_\Omega L_i(\bm{l}) (\bm{l} \cdot \bm{n}) d\bm{l} +  \int_\Omega  L_i(\bm{l}) f^\text{s}_r(\bm{l}, \bm{v}) (\bm{l} \cdot \bm{n}) d\bm{l} \\
\approx& f^\text{d}_r I^\text{d} + R I^\text{s} \\
=& \hat{\bm{c}}^\text{d}_\text{def}(\bm{u}) + \hat{\bm{c}}^\text{s}_\text{def}(\bm{u}).
\end{aligned}
\end{equation}

\begin{table*}[!t]
\centering
\caption{Quantatitive Comparisons on object-level datasets (\ie TensoIR, Shiny Blender, and Glossy Blender datasets).
The methods in the second block support relighting applications, and the methods marked with $\dag$ achieve real-time rendering.
Our method achieves the best results on TensoIR and Glossy Blender datasets.
}
\begin{tabular}{@{}lc c c cc c c cc c c}
\toprule
\multirow{2}{*}{Method} &
\multicolumn{4}{c}{TensoIR \cite{jin2023tensoir}} &
\multicolumn{4}{c}{Shiny Blender \cite{verbin2022ref}} &
\multicolumn{3}{c}{Glossy Blender \cite{liu2023nero}} \\
\cmidrule(lr){2-5}\cmidrule(lr){6-9}\cmidrule(lr){10-12}
&
PSNR $\uparrow$ & SSIM $\uparrow$ & LPIPS $\downarrow$ & MAE $\downarrow$ &
PSNR $\uparrow$ & SSIM $\uparrow$ & LPIPS $\downarrow$ & MAE $\downarrow$ &
PSNR $\uparrow$ & SSIM $\uparrow$ & LPIPS $\downarrow$ \\
\midrule
Ref-NeRF~\cite{verbin2022ref} &
37.75 & 0.982 & 0.026 & 7.585 &
35.96 & 0.967 & 0.059 & 13.94 &
27.86 & 0.878 & 0.114 \\
3DGS$^\dag$~\cite{kerbl20233d} &
40.82 & 0.990 & 0.014 & - &
30.37 & 0.947 & 0.083 & - &
24.96 & 0.872 & 0.104 \\
2DGS$^\dag$~\cite{huang20242d} &
39.23 & 0.987 & 0.019 & 4.165 &
29.47 & 0.946 & 0.084 & 13.647 &
26.00 & 0.915 & 0.087 \\
\midrule
NDR$^\dag$~\cite{munkberg2022extracting} &
30.19 & 0.962 & 0.052 & 6.079 &
26.36 & 0.886 & 0.261 & 17.047 &
22.11 & 0.856 & 0.165 \\
NeRO~\cite{liu2023nero} &
29.08 & 0.949 & 0.067 & 4.969 &
29.84 & \cc{3}0.962 & 0.072 & \cc{2}2.360 &
\cc{3}28.76 & \cc{2}0.943 & \cc{2}0.062 \\
ENVIDR~\cite{liang2023envidr} &
33.37 & 0.961 & 0.049 & 8.027 &
\cc{1}35.85 & \cc{1}0.983 & \cc{1}0.036 & \cc{3}4.608 &
\cc{2}29.23 & \cc{2}0.943 & \cc{3}0.065 \\
TensoIR~\cite{jin2023tensoir} &
35.09 & 0.976 & 0.041 & \cc{1}4.100 &
29.56 & 0.927 & 0.130 & 5.408 &
24.85 & 0.872 & 0.149 \\
RelightGS~\cite{gao2023relightable} &
\cc{2}37.57 & \cc{3}0.983 & \cc{1}0.020 & 6.078 &
27.31 & 0.920 & 0.120 & 6.115 &
24.21 & 0.893 & 0.106 \\
GS-IR$^\dag$~\cite{liang2024gs} &
35.33 & 0.974 & 0.039 & \cc{3}4.947 &
26.88 & 0.887 & 0.141 & 9.464 &
23.58 & 0.813 & 0.168 \\
GShader$^\dag$~\cite{jiang2024gaussianshader} &
\cc{3}37.54 & \cc{2}0.984 & \cc{3}0.022 & 6.525 &
\cc{3}31.94 & 0.957 & \cc{3}0.068 & 6.115 &
26.04 & 0.911 & 0.104 \\
Ours$^\dag$ &
\cc{1}37.61 & \cc{1}0.985 & \cc{1}0.020 & \cc{2}4.485 & 
\cc{2}34.28 & \cc{2}0.973 & \cc{2}0.056 & \cc{1}2.193 &
\cc{1}29.44 & \cc{1}0.953 & \cc{1}0.057 \\
\bottomrule
\end{tabular}
\label{tab:compairson_object}
\end{table*}


In a nutshell, the decomposed results of forward and deferred shading schemes are:
\begin{equation}
\label{eq:shading_results2}
\begin{aligned}
\hat{\bm{c}}^\text{d}_\text{for}(\bm{u}) =& f^\text{d}_r \sum^N_{k=1}  w_k I^\text{d}_\text{k},\quad&
\hat{\bm{c}}^\text{s}_\text{for}(\bm{u}) =& \sum^N_{k=1}w_k R_k I^\text{s}_k, \\
\hat{\bm{c}}^\text{d}_\text{def}(\bm{u}) =& f^\text{d}_r I^\text{d},\quad&
\hat{\bm{c}}^\text{s}_\text{def}(\bm{u}) =& R I^\text{s}.
\end{aligned}
\end{equation}

\cref{fig:forward_deferred_double} demonstrates more intuitive comparisons.
Even though we assume all particles have the same material, different particles covering the pixel $\bm{u}$ have various normals $\{\bm{n}_k\}^N_{k=1}$, resulting in various upper hemispheres $\{\Omega_k\}^N_{k=1}$ and BRDF lobes $\{\Phi_k\}^N_{k=1}$.
Thus the integration domain of incidence in forward shading contributing to the results indeed exceeds a hemisphere and a lobe in terms of diffuse and specular terms, which means that the forward shading scheme tends to produce fuzzy results and achieves better results than the deferred shading scheme in scenes featuring rough surfaces.
However, the highlights on the glossy surface are almost contributed by the incidence within extremely narrow lobes according to the importance sampling principle~\cite{veach1998robust}.
Therefore, under the premise that all materials are constant, the forward shading scheme further requires ensuring all particles' upper hemispheres (\ie normals) are constant to achieve identical rendering results, which is impractical.
The forward shading scheme is weak at handling glossy surfaces, which is the strength of the deferred shading scheme.
Our experiments in \cref{subsec:ablation} verify this analysis.
In addition, the deferred shading scheme supports modifications on the G-Buffer to change the rendering results, which is friendly to the editing task.

Based on it, we are motivated to propose shading unification in GUS-IR for inverse rendering.
As shown in \cref{fig:pipeline}, GUS-IR consists of forward and deferred shading branches.
During training, we perform both branches simultaneously and employ the same supervision.
Benefitting from this combination of two branches, GUS-IR achieves remarkable results in most complex scenes, whether the scenes mainly exist on rough or glossy surfaces.
In the inference, we only maintain the deferred shading branch.

\subsection{Occlusion Volumes}
\label{subsec:occlusion}
Modeling occlusion is essential to handle indirect illumination.
Ambient occlusion (AO) represents the accessibility of a surface point, and is an efficient and promising way to tackle the low-frequent occlusion.
To model the AO in a scene represented by particles in 3DGS, as mentioned at~\cref{sec:preliminary}, GS-IR uses spherical harmonics (SH) bases to convolve the binary occlusion cubemap $\{\bm{D}^{\bm{v}^\text{O}}_i\}^6_{i=1}$ and obtain SH coefficients $\bm{f}^\text{O}$ to cache low-frequency occlusion in regularly placed volumes.
However, the SH convolution cannot perfectly approximate the result of the hemispherical integration of the occlusion cubemap and suffers from the `leaking` problem.
To this end, in GUS-IR, we decided to directly store the binary occlusion cubemap in each volume and obtain AO through runtime sampling.
As shown in \cref{fig:baking}, we follow GS-IR, regularly place the volumes in the bounded 3D scene, and get the occlusion cubemap $\{\bm{D}^{\bm{v}^\text{O}}_i\}^6_{i=1}$ in the same way.
In contrast to GS-IR which compresses the occlusion information in a set of SH coefficients, we flatten the occlusion cubemap into a dense byte array, where each byte can represent 8 pixels' occlusion.

In the decomposition stage, we follow GS-IR and conduct normal-aware masked-trilinear interpolation to get the occlusion cubemap for each particle.
For a particle with position $\bm{x}$ and normal $\bm{n}$, we uniform sample directions of its upper hemisphere centered at $\bm{n}$ and query the occlusion on this array based on the directions.
Then we average the values as the particle's ambient occlusion $O(\bm{x})$.
The ambient occlusion $O(\bm{x})$ is used in \cref{eq:diffuse}.
For deferred shading, we conduct volumetric accumulation to obtain the pixel's ambient occlusion.
The whole recovery still achieves real-time performance.

\begin{figure*}[tb]
\centering
\includegraphics[width=\linewidth]{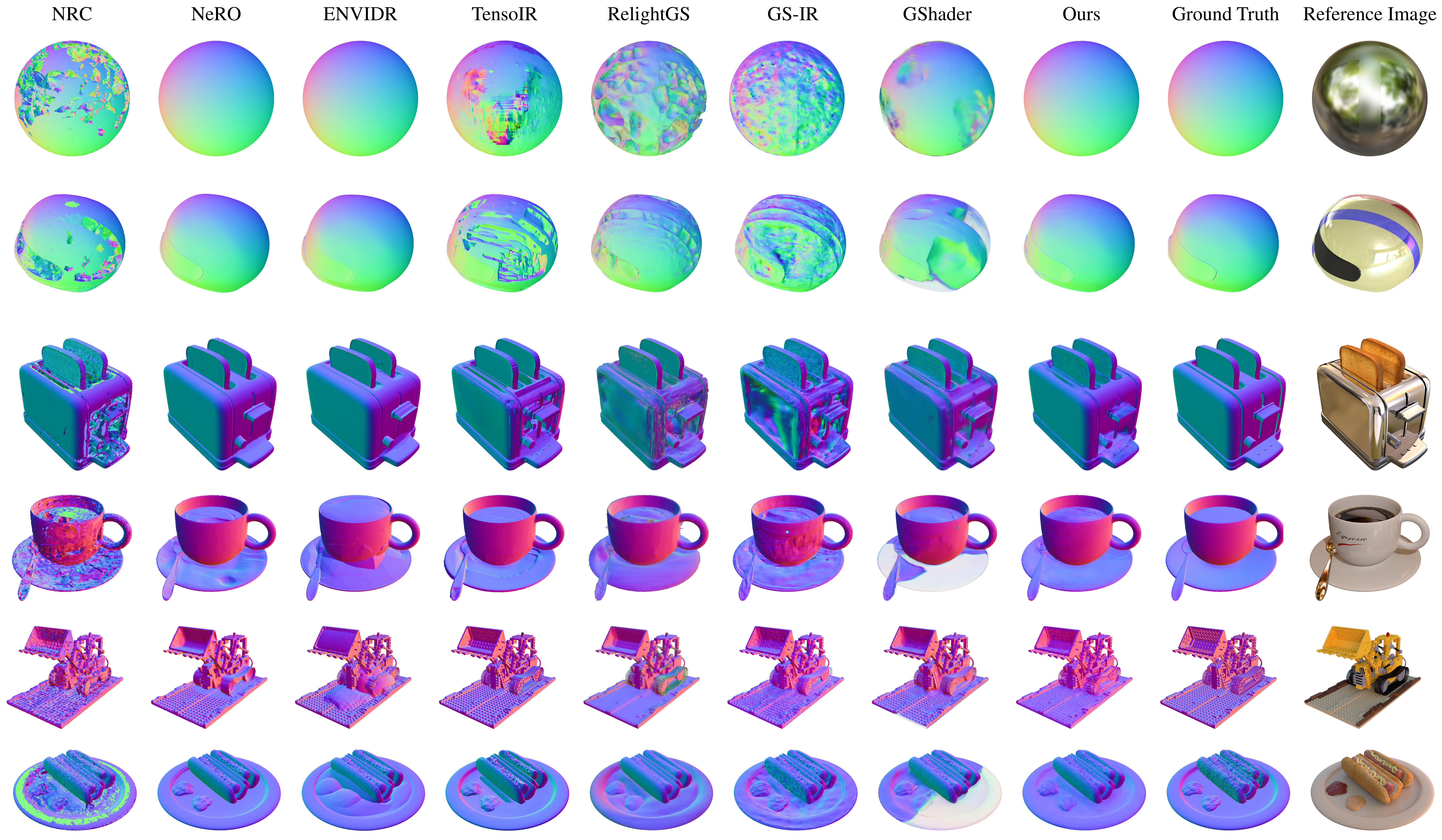}
\caption{Normal estimation comparisons on TensoIR~\cite{jin2023tensoir} and Shiny Blender~\cite{verbin2022ref} datasets. We exemplify the normal estimation results of our GUS-IR and other cutting-edge methods.}
\label{fig:normal_comparison}
\end{figure*}

\begin{figure*}[tb]
\centering
\includegraphics[width=\linewidth]{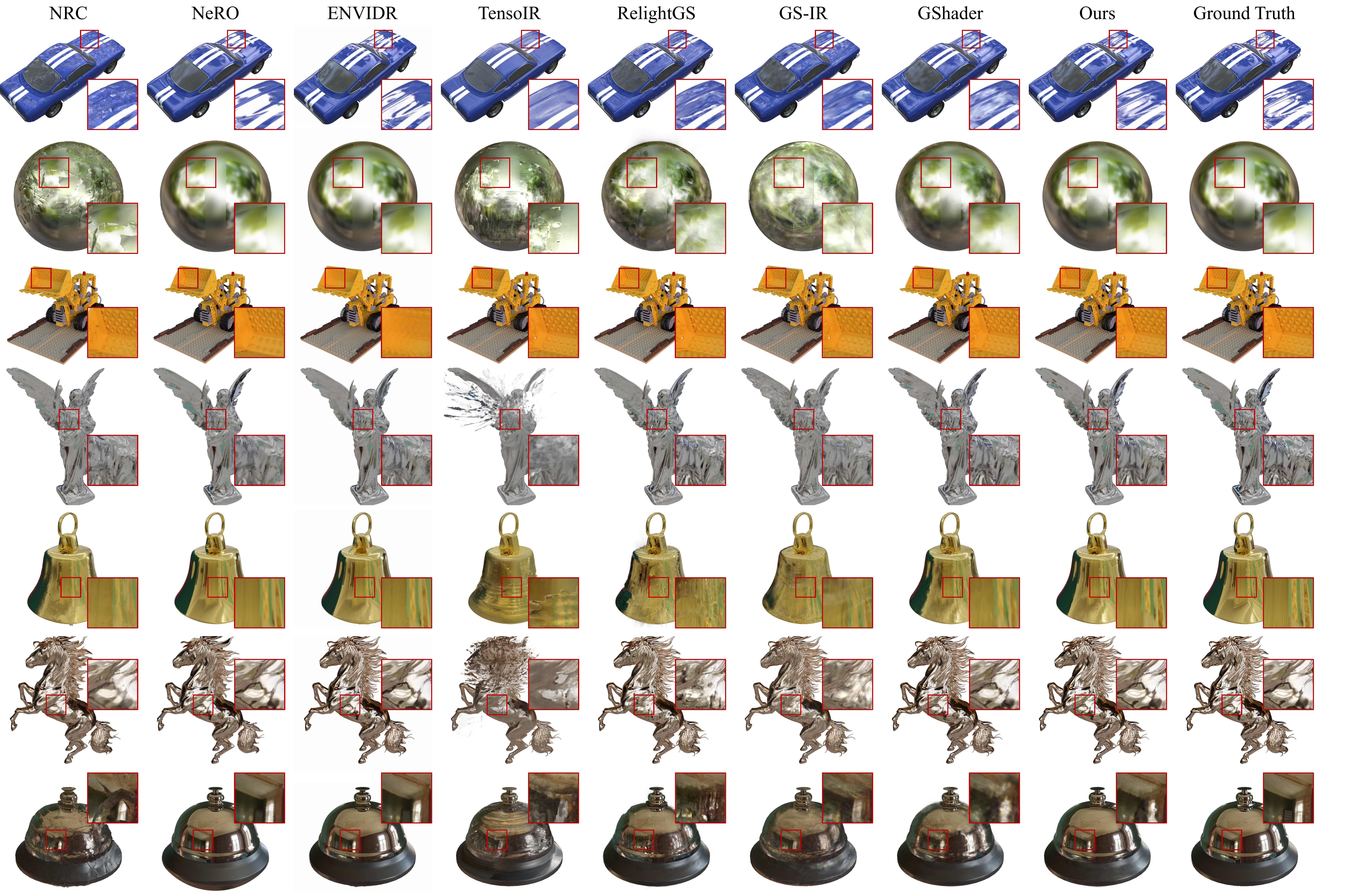}
\caption{Novel-view synthesis comparisons on object-level datasets (\ie TensoIR, Shiny Blender, and Glossy Blender datasets). We exemplify the rendering results of our GUS-IR and other cutting-edge methods.
}
\label{fig:nvs_comparison}
\end{figure*}

\subsection{Losses}
\label{subsec:losses}
In GUS-IR, we simultaneously supervise the forward and deferred shading results in \cref{eq:forward_approx} and \cref{eq:deferred_approx}:
\begin{equation}
\label{eq:loss_forward_deferred}
\begin{aligned}
\mathcal{L}_\text{f} &= (1 - \lambda) \mathcal{L}_1(\hat{\bm{C}}_\text{for}, \bm{C}) +
\lambda \mathcal{L}_\text{D-SSIM}(\hat{\bm{C}}_\text{for}, \bm{C}), \\
\mathcal{L}_\text{d} &= (1 - \lambda) \mathcal{L}_1(\hat{\bm{C}}_\text{def}, \bm{C}) +
\lambda \mathcal{L}_\text{D-SSIM}(\hat{\bm{C}}_\text{def}, \bm{C}),
\end{aligned}
\end{equation}
where $\bm{C}, \hat{\bm{C}}_\text{for}, \hat{\bm{C}}_\text{def}$ denote the reference image, the images rendered by performing forward and deferred shading schemes, respectively.

Following GS-IR~\cite{liang2024gs} and GaussianShader~\cite{jiang2024gaussianshader}, we also and leverage the gradient derived from the rendered depth map to supervise the learned normal and encourage the volumetric-accumulated alpha to approach 0 or 1:
\begin{equation}
\label{eq:loss_reg}
\begin{aligned}
\mathcal{L}_\text{n} =& \frac{1}{\Vert \bm{C} \Vert}\sum_{\bm{u} \subset \bm{C}}\left[
1 - \hat{\bm{n}}(\bm{u})^\top \cdot \nabla \hat{d}(\bm{u})
\right], \\
\mathcal{L}_\alpha =& \frac{1}{\Vert \bm{C} \Vert}\sum_{\bm{u} \subset \bm{C}} \left[
\log(\alpha(\bm{u})) + \log(1 - \alpha(\bm{u}))
\right]
, \\
\mathcal{L}_\text{reg} =& \lambda_\text{n} \mathcal{L}_\text{n} + \lambda_\alpha \mathcal{L}_\alpha,
\end{aligned}
\end{equation}
where $\hat{\bm{n}}(\bm{u})$ denotes the volumetric accumulated normal of pixel $\bm{u}$, $\nabla \hat{d}(\bm{u})$ denotes the gradient to the depth of pixel $\bm{u}$, $\lambda_n = 0.1$ and $\lambda_\alpha = 0.001$.
The summary loss $\mathcal{L}$ is:
\begin{equation}
\label{eq:losses}
\begin{aligned}
\mathcal{L} =& \mathcal{L}_\text{f} + \mathcal{L}_\text{d} + \mathcal{L}_\text{reg}.
\end{aligned}
\end{equation}

\begin{figure*}[tb]
\centering
\includegraphics[width=1.\linewidth]{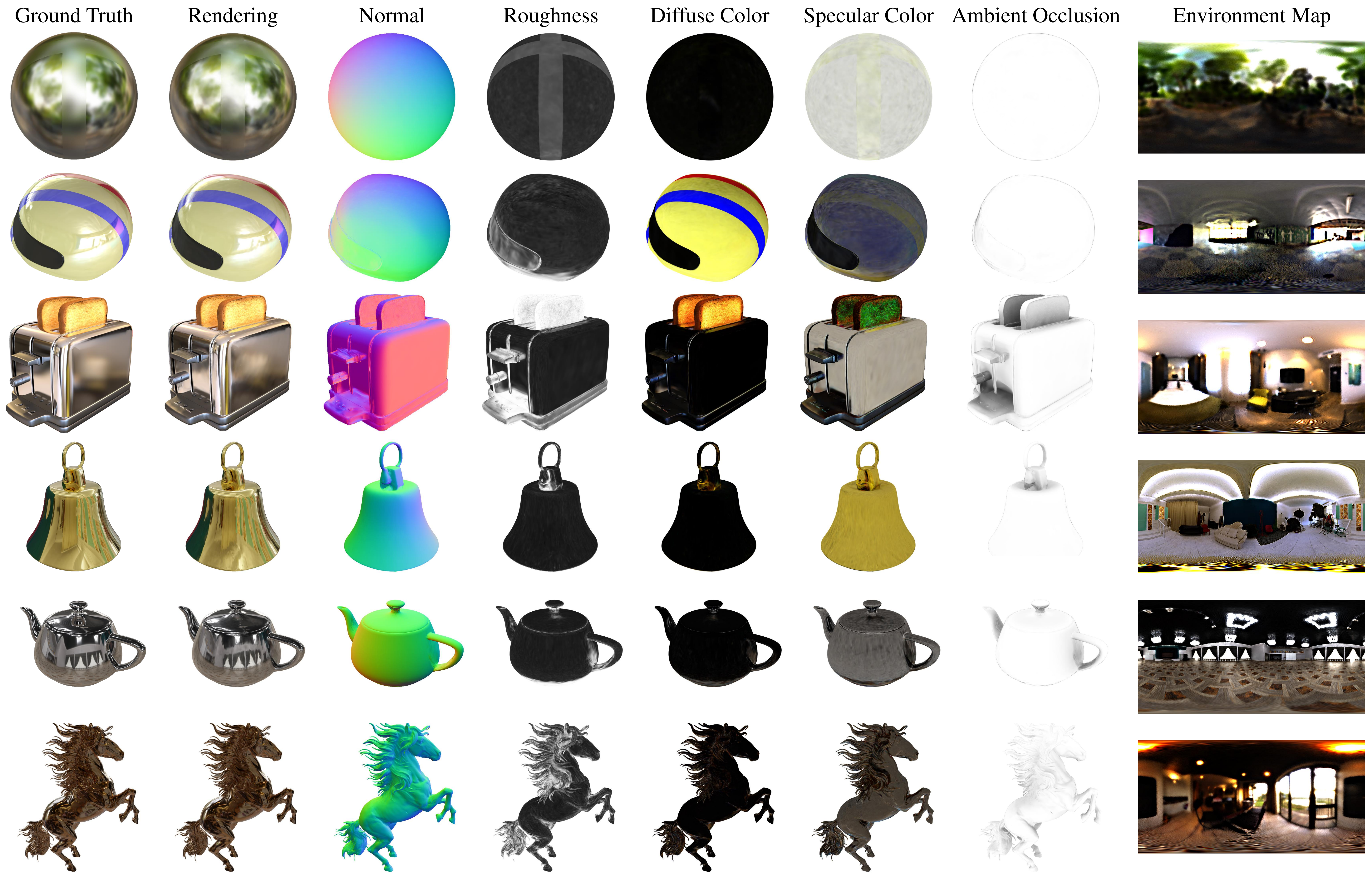}
\caption{Decomposition results of GUS-IR on Shiny Blender~\cite{verbin2022ref} and Glossy Blender~\cite{liu2023nero} datasets, including objects with glossy surfaces.}
\label{fig:glossy_decomposition}
\end{figure*}

\begin{table}[t]
\centering
\caption{Quantatitive Comparisons of Relighiting on TensoIR and Glossy Blender datasets. The methods marked with $\dag$ achieve real-time rendering.}
\scalebox{0.95}{
\setlength{\tabcolsep}{3pt}
\begin{tabular}{@{}l c c c c c c}
\toprule
\multirow{2}{*}{Method} &
\multicolumn{3}{c}{TensoIR~\cite{jin2023tensoir}} &
\multicolumn{3}{c}{Glossy Blender~\cite{liu2023nero}} \\
\cmidrule(lr){2-4}\cmidrule(lr){5-7}
&
PSNR $\uparrow$ & SSIM $\uparrow$ & LPIPS $\downarrow$ &
PSNR $\uparrow$ & SSIM $\uparrow$ & LPIPS $\downarrow$ \\
\midrule
NeRO~\cite{liu2023nero} &
\cc{3}26.20 & \cc{3}0.891 & 0.096 &
\cc{1}28.77 & \cc{1}0.950 & \cc{1}0.057 \\
ENVIDR~\cite{liang2023envidr} &
24.81 & 0.869 & \cc{3}0.095 &
19.21 & 0.789 & 0.115 \\
TensoIR~\cite{jin2023tensoir} &
\cc{1}28.58 & \cc{1}0.944 & \cc{1}0.081 &
15.19 & 0.786 & 0.213 \\
RelightGS~\cite{gao2023relightable} &
24.41 & 0.890 & 0.100 &
20.22 & 0.856 & 0.112 \\
NDR$^\dag$~\cite{munkberg2022extracting} &
19.88 & 0.879 & 0.104 &
\cc{3}24.54 & 0.859 & 0.135 \\
GS-IR$^\dag$~\cite{liang2024gs} &
24.37 & 0.885 & 0.096 &
17.50 & 0.818 & 0.136 \\
GShader$^\dag$~\cite{liang2024gs} &
22.42 & 0.872 & 0.103 &
21.61 & \cc{3}0.874 & \cc{3}0.105 \\
Ours$^\dag$ &
\cc{2}27.53 & \cc{2}0.919 & \cc{2}0.085 &
\cc{2}25.17 & \cc{2}0.922 & \cc{2}0.069 \\
\bottomrule
\end{tabular}
}
\label{tab:relighting}
\end{table}

\begin{table*}[t]
\centering
\caption{Quantatitive Comparisons on scene-level datasets (\ie Mip-NeRF 360 and Ref-NeRF Real datasets). 
The methods in the second block support relighting applications, and the methods marked with $\dag$ achieve real-time rendering.
Please note that our GUS-IR achieves comparable performance on the Ref-NeRF Real dataset against other methods that only focus on novel-view synthesis.}
\setlength{\tabcolsep}{10pt}
\begin{tabular}{@{}l c c c c c c c c c}
\toprule
\multirow{2}{*}{Method} &
\multicolumn{3}{c}{Mip-NeRF 360 Outdoor~\cite{barron2022mip}} &
\multicolumn{3}{c}{Mip-NeRF 360 Indoor~\cite{barron2022mip}} &
\multicolumn{3}{c}{Ref-NeRF Real~\cite{verbin2022ref}} \\
\cmidrule(lr){2-4}\cmidrule(lr){5-7}\cmidrule(lr){8-10}
&
PSNR $\uparrow$ & SSIM $\uparrow$ & LPIPS $\downarrow$ &
PSNR $\uparrow$ & SSIM $\uparrow$ & LPIPS $\downarrow$ &
PSNR $\uparrow$ & SSIM $\uparrow$ & LPIPS $\downarrow$ \\
\midrule
Mip-NeRF 360~\cite{barron2022mip} &
24.43 & 0.694 & 0.278 &
31.49 & 0.918 & 0.179 &
24.27 & 0.649 & 0.276 \\
UniSDF~\cite{wang2023unisdf} &
24.77 & 0.723 & 0.241 &
31.28 & 0.901 & 0.181 &
23.70 & 0.635 & 0.266 \\
3DGS$^\dag$~\cite{kerbl20233d} &
24.64 & 0.731 & 0.234 &
30.41 & 0.917 & 0.190 &
23.67 & 0.632 & 0.288 \\
2DGS$^\dag$~\cite{huang20242d} &
24.33 & 0.709 & 0.284 &
30.39 & 0.922 & 0.183 &
23.65 & 0.634 & 0.285 \\
\midrule
GS-IR$^\dag$~\cite{liang2024gs} &
23.45 & 0.671 & 0.284 &
27.80 & 0.870 & 0.248 &
23.32 & 0.625 & 0.283 \\
GaussianShader$^\dag$~\cite{jin2023tensoir} &
22.80 & 0.665 & 0.297 &
26.61 & 0.878 & 0.243 &
22.96 & 0.624 & 0.294 \\
Ours$^\dag$ &
\textbf{23.76} & \textbf{0.696} & \textbf{0.276} &
\textbf{28.98} & \textbf{0.902} & \textbf{0.222} &
\textbf{23.67} & \textbf{0.646} & \textbf{0.282} \\
\bottomrule
\end{tabular}
\label{tab:comparison_scene}
\end{table*}

\section{Experiments}
\label{sec:experiments}
\subsection{Implementation}
\noindent\textbf{Dataset \& Metrics}
We conduct experiments on three object-level datasets (\ie TensorIR Synthetic~\cite{jin2023tensoir}, Shiny Blender~\cite{verbin2022ref} datasets, and Glossy Blender~\cite{liu2023nero} datasets) and two real scene-level datasets (\ie Mip-NeRF 360~\cite{barron2022mip} and Ref-NeRF Real~\cite{verbin2022ref} datasets), which contain $18$ objects and $12$ publicly available scenes, covering both rough and glossy surfaces.
Specifically, TensoIR and Shiny Blender datasets provide normal references for evaluating the quality of geometry reconstruction.
TensoIR and Glossy Blender datasets further support the relighting evaluation.
Generally, we evaluate the novel-view synthesis results on all datasets in terms of Peak Signal-to-Noise Ratio (PSNR), Structural Similarity Index Measure (SSIM), and Learned Perceptual Image Patch Similarity (LPIPS)~\cite{zhang2018unreasonable}.
To evaluate the relighting results, we use the same metrics as novel-view synthesis.
In addition, we leverage mean angular error (MAE) to evaluate the normal estimation quality.

\vspace{6pt}\noindent\textbf{Implementation Detilas}
In the first stage, we follow the training schedule of 3DGS~\cite{kerbl20233d} to optimize the GUS-IR for 30K iterations using the Adam optimizer~\cite{kingma2014adam} and use the loss defined in \cref{eq:losses} for supervision.
For caching the occlusion in the baking stage, we regularly place $128^3$ volumes in the space $[-1.5, 1.5]^3$ for objects and $256^3$ volumes in the space $[-8, 8]^3$ for scenes to cache the occlusion and model the indirect irradiance.
After the baking stage, we only use $\mathcal{L}_\text{f}$ and $\mathcal{L}_\text{d}$ defined in \cref{eq:loss_forward_deferred} to optimize GUS-IR for 5K iterations.

\subsection{Comparisons}
\label{subsec:comparisons}
We conduct a comprehensive comparison against cutting-edge novel-view synthesis and inverse rendering methods on public datasets.
All methods take multi-view images captured under unknown lighting conditions as input.

\begin{figure*}[tb]
\centering
\includegraphics[width=1.\linewidth]{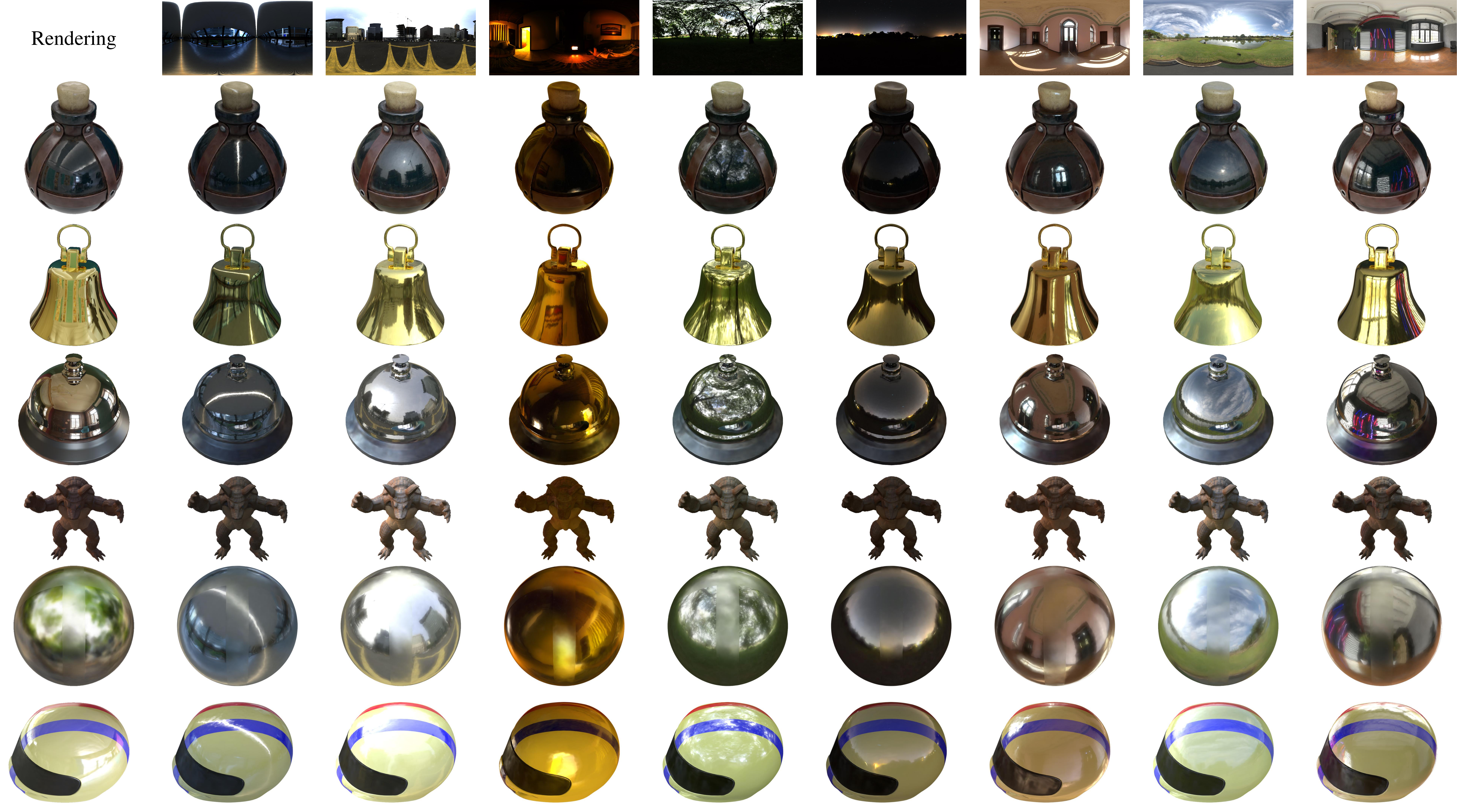}
\caption{Relighting results on object-level datasets.
We perform relighting experiments on objects using the geometry, and material recovered from our GUS-IR method.
We test our method under different lighting conditions.
The results show that our method handles both rough and glossy surfaces well.}
\label{fig:relighting_object}
\end{figure*}

\vspace{6pt}\noindent\textbf{Object-level}
In \cref{tab:compairson_object}, we show the novel-view synthesis and normal estimation performance of all methods on the object-level datasets (\ie TensoIR~\cite{jin2023tensoir}, Shiny Blender~\cite{verbin2022ref}, and Glossy Blender~\cite{liu2023nero}).
Our methods almost surpass other relightable methods.
The qualitative comparisons of normal estimation and novel-view synthesis are respectively shown in \cref{fig:normal_comparison} and \cref{fig:nvs_comparison}.
It is worth noting that these datasets contain complex rough and glossy surfaces.
Both quantitative and qualitative results demonstrate that our method can effectively decompose any surface.
Notably, GUS-IR is one of the few methods that achieve real-time rendering.
In addition, our method also achieves remarkable relighting performance as shown in \cref{tab:relighting}.
\cref{fig:relighting_object} shows the relighting results of GUS-IR on objects.

\begin{figure*}[!t]
\centering
\includegraphics[width=1.0\linewidth]{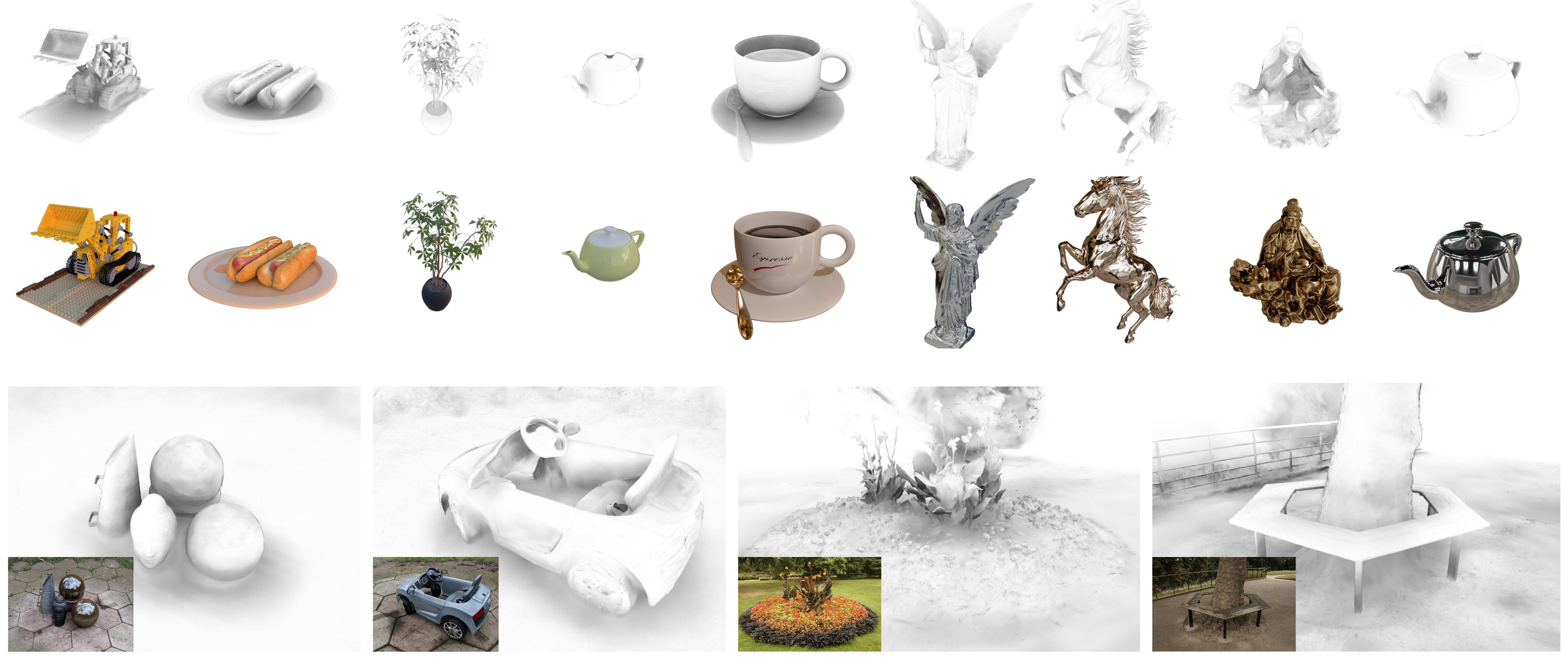}
\caption{Visualization of the recovered ambient occlusion. The visualization highlights the intricate shadowing and occlusion details captured by our GUS-IR method, emphasizing the performance of our approach in modeling indirect illumination.}
\label{fig:occlusion}
\end{figure*}

\begin{figure*}[!t]
\centering
\includegraphics[width=1.0\linewidth]{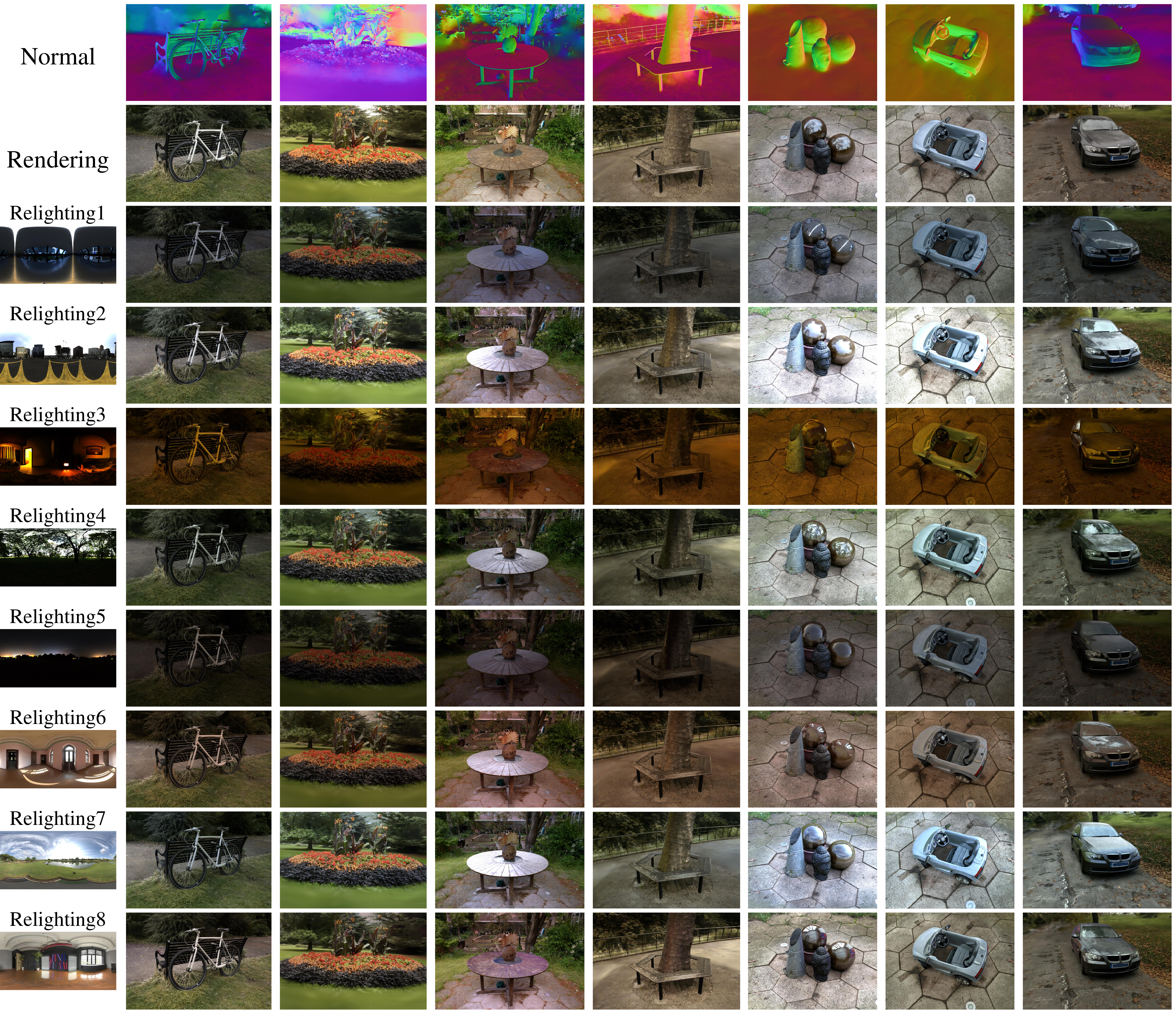}
\caption{Relighting results on scene-level datasets. We perform relighting experiments on scenes using the geometry, and material recovered from our GUS-IR method.
The results in the second and third columns on the right show that GUS-IR resolves the glossy marble surfaces and the car metal shell, reflecting clear environment illumination.
}
\label{fig:relighting_scene}
\end{figure*}

\vspace{6pt}\noindent\textbf{Scene-level}
In \cref{tab:comparison_scene}, we show the novel-view synthesis comparisons on the scene-level datasets, including Mip-NeRF 360~\cite{barron2022mip} and Ref-NeRF Real~\cite{verbin2022ref} datasets.
The results show that GUS-IR achieves the best performance among relightable methods.
Notably, GUS-IR achieves comparable performance to other methods that only focus on novel-view synthesis.
\cref{fig:relighting_scene} shows the relighting results of GUS-IR on scenes.

In summary, GUS-IR achieves high-quality real-time inverse rendering by incorporating unified shading into the 3DGS framework, which effectively decomposes the physical properties of rough and glossy surfaces in complex scenes.
We further show the decomposition on the glossy surfaces of GUS-IR in \cref{fig:glossy_decomposition}.
The relighting results in \cref{fig:relighting_object} and \cref{fig:relighting_scene} verify that GUS-IR can reasonably decompose physical properties and produce harmonious relighting effects.

\begin{table*}
\centering
\caption{
Ablation on object-level datasets (\ie TensoIR~\cite{jin2023tensoir}, Shiny Blender~\cite{verbin2022ref}, and Glossy Blender~\cite{liu2023nero} datasets).
\textit{Only Forward} and \textit{Only Deferred} respectively refer to only employing forward or deferred shading schemes, Other alternatives use the unified shading scheme.
\textit{Attach Normal} indicates attaching a learnable normal for each particle.
\textit{Residual Normal} denotes associating a predicted residual on the shortest axis as the normal for each particle.
\textit{W/o Indirect} denotes ignoring ambient occlusion and indirect illumination.
}
\begin{tabular}{@{}lc c c c c c c c c c c}
\toprule
\multirow{2}{*}{Method} &
\multicolumn{4}{c}{TensoIR \cite{jin2023tensoir}} &
\multicolumn{4}{c}{Shiny Blender \cite{verbin2022ref}} &
\multicolumn{3}{c}{Glossy Blender \cite{liu2023nero}} \\
\cmidrule(lr){2-5}\cmidrule(lr){6-9}\cmidrule(lr){10-12}
&
PSNR $\uparrow$ & SSIM $\uparrow$ & LPIPS $\downarrow$ & MAE $\downarrow$ &
PSNR $\uparrow$ & SSIM $\uparrow$ & LPIPS $\downarrow$ & MAE $\downarrow$ &
PSNR $\uparrow$ & SSIM $\uparrow$ & LPIPS $\downarrow$ \\
\midrule
Only Forward &
\textbf{37.74} & \textbf{0.985} & \textbf{0.020} & 4.680 &
33.15 & 0.965 & 0.067 & 2.436 &
28.64 & 0.928 & 0.084 \\
Only Deferred &
36.35 & 0.980 & 0.029 & 4.769 &
33.58 & 0.970 & 0.059 & 2.370 &
29.12 & 0.950 & 0.059 \\
Attach Normal &
37.67 & 0.984 & 0.022 & 5.433 &
33.43 & 0.968 & 0.060 & 2.457 &
28.89 & 0.946 & 0.063 \\
Residual Normal &
37.62 & \textbf{0.985} & 0.021 & 4.761 &
34.17 & 0.971 & 0.058 & 2.221 &
29.43 & 0.952 & 0.063 \\
W/o Indirect &
37.37 & 0.983 & 0.024 & \textbf{4.485} &
34.20 & 0.970 & 0.058 & \textbf{2.193} &
29.28 & 0.951 & 0.058 \\
Ours &
37.61 & \textbf{0.985} & \textbf{0.020} & \textbf{4.485} & 
\textbf{34.28} & \textbf{0.973} & \textbf{0.056} & \textbf{2.193} &
\textbf{29.44} & \textbf{0.953} & \textbf{0.057} \\
\bottomrule
\end{tabular}
\label{tab:object_ablation}
\end{table*}

\begin{table}[!t]
\centering
\caption{Analysis of the impact of different conditions to relighting on object-level datasets (\ie TensoIR~\cite{jin2023tensoir} and Glossy Blender~\cite{verbin2022ref} datasets).}
\scalebox{0.93}{
\setlength{\tabcolsep}{3pt}
\begin{tabular}{@{}l c c c c c c}
\toprule
\multirow{2}{*}{Method} &
\multicolumn{3}{c}{TensoIR~\cite{jin2023tensoir}} &
\multicolumn{3}{c}{Glossy Blender~\cite{liu2023nero}} \\
\cmidrule(lr){2-4}\cmidrule(lr){5-7}
&
PSNR $\uparrow$ & SSIM $\uparrow$ & LPIPS $\downarrow$ &
PSNR $\uparrow$ & SSIM $\uparrow$ & LPIPS $\downarrow$ \\
\midrule
Only Forward &
27.15 & 0.917 & 0.086 & 
23.16 & 0.873 & 0.105 \\
Only Defer &
26.83 & 0.915 & 0.087 & 
23.99 & 0.903 & 0.073 \\
Attach Normal &
26.12 & 0.913 & 0.089 & 
22.81 & 0.881 & 0.092 \\
Residual Normal &
26.32 & 0.914 & 0.089 & 
24.72 & 0.911 & 0.073 \\
Ours &
\textbf{27.53} & \textbf{0.919} & \textbf{0.085} &
\textbf{25.17} & \textbf{0.922} & \textbf{0.069} \\
\bottomrule
\end{tabular}
}
\label{tab:ablation_relighting}
\end{table}

\begin{table*}
\centering
\caption{
Ablation on scene-level datasets (\ie Mip-NeRF 360~\cite{barron2022mip} and Ref-NeRF Real~\cite{verbin2022ref} datasets).
\textit{Only Forward} and \textit{Only Deferred} respectively refer to only employing forward or deferred shading schemes, Other alternatives use the unified shading scheme.
\textit{Attach Normal} indicates attaching a learnable normal for each particle.
\textit{Residual Normal} denotes associating a predicted residual on the shortest axis as the normal for each particle.
\textit{W/o Indirect} denotes ignoring ambient occlusion and indirect illumination.
}
\setlength{\tabcolsep}{10pt}
\begin{tabular}{@{}l c c c c c c c c c}
\toprule
\multirow{2}{*}{Method} &
\multicolumn{3}{c}{Mip-NeRF 360 Outdoor~\cite{barron2022mip}} &
\multicolumn{3}{c}{Mip-NeRF 360 Indoor~\cite{barron2022mip}} &
\multicolumn{3}{c}{Ref-NeRF Real~\cite{verbin2022ref}} \\
\cmidrule(lr){2-4}\cmidrule(lr){5-7}\cmidrule(lr){8-10}
&
PSNR $\uparrow$ & SSIM $\uparrow$ & LPIPS $\downarrow$ &
PSNR $\uparrow$ & SSIM $\uparrow$ & LPIPS $\downarrow$ &
PSNR $\uparrow$ & SSIM $\uparrow$ & LPIPS $\downarrow$ \\
\midrule
Only Forward &
23.72 & 0.695 & 0.281 &
\textbf{29.08} & \textbf{0.904} & 0.223 & 
22.89 & 0.619 & 0.305 \\
Only Defer &
23.53 & 0.679 & 0.285 &
28.84 & 0.891 & 0.233 & 
23.49 & 0.641 & \textbf{0.278} \\
Attach Normal &
\textbf{23.89} & \textbf{0.701} & \textbf{0.269} &
28.93 & 0.902 & 0.225 & 
23.02 & 0.635 & 0.282 \\
Residual Normal &
23.79 & 0.699 & 0.273 &
28.90 & 0.901 & 0.223 & 
23.66 & 0.648 & 0.280 \\
W/o Indirect &
23.66 & 0.693 & 0.280 & 
28.52 & 0.900 & 0.225 & 
23.56 & 0.642 & 0.285 \\
Ours &
23.76 & 0.696 & 0.276 &
28.98 & 0.902 & \textbf{0.222} &
\textbf{23.67} & \textbf{0.646} & 0.282 \\
\bottomrule
\end{tabular}
\label{tab:scene_ablation}
\end{table*}

\subsection{Ablation Studies}
\label{subsec:ablation}
To evaluate the efficacy of our proposed schemes, we design elaborate experiments on the same five datasets as in the above comparison for ablation.
We analyze the impact of different shading schemes, normal representations, and modeling indirect illumination on the final performance.

\vspace{6pt}\noindent\textbf{Analysis on Different Shading Schemes}
Employing a correct shading scheme is critical to inverse rendering, as it relates to the decomposition quality.
As stated in \cref{subsec:unification}, our employed unified shading scheme not only restores the diffuse material of rough surfaces but also captures the highlight details, handling glossy surfaces.
The results in \cref{tab:object_ablation} and \cref{tab:scene_ablation} show that the forward shading slightly outperforms other schemes on the TensoIR and Mip-NeRF 360 indoor datasets that are full of rough surfaces.
However, it is significantly inferior to other schemes on the Shiny Blender and Glossy Blender datasets, mainly containing objects with glossy surfaces. 
The relighting results on \cref{tab:ablation_relighting} further verify the efficacy of our scheme.

\vspace{6pt}\noindent\textbf{Analysis on Normal Representations}
Modeling normals is the primary problem faced when using 3DGS for 3D reconstruction and inverse rendering.
The results in \cref{tab:object_ablation}, \cref{tab:scene_ablation}, and \cref{tab:ablation_relighting} show that our employed scheme (using the shortest axis as the normal for each particle) is almost superior to \textit{Attach Normal} and \textit{Residual Normal} in novel-view synthesis, especially in relighting tasks.
It is worth noting that the normal estimation (\ie MAE) of our employed scheme significantly outperforms \textit{Attach Normal} and \textit{Residual Normal}, and the \textit{Residual Normal} surpasses \textit{Attach Normal Normal}.

\vspace{6pt}\noindent\textbf{Analysis on Indirect Illumination}
To verify the efficacy of modeling indirect illumination, we compare our results with those of the first stage, skipping baking and subsequent stages, referred to \textit{W/o Indirect}.
Results in 
\cref{tab:object_ablation}, \cref{tab:scene_ablation} verify that modeling diffuse indirect illumination using our proposed scheme as illustrated at \cref{subsec:occlusion} improves the novel-view synthesis results.
\cref{fig:occlusion} showcases the ambient occlusion produced by our proposed scheme, which verifies that our scheme can capture the complex shadowing and occlusion.

\section{Conclusion}
\label{sec:conclusion}
In this paper, we introduce a new GUS-IR framework for recovering Gaussians with Unified Shading for Inverse Rendering from a set of images with natural illumination. Our approach achieves impressive intrinsic decomposition and relighting on both glossy objects and complex scenes. We conduct a thorough comparison and analysis of two prominent shading schemes, namely forward shading and deferred shading, commonly used for 3DGS-based inverse rendering tasks. Based on this analysis, we propose a unified shading solution that combines the advantages of both shading schemes, leading to better intrinsic decomposition suitable for complex scenes with rough surfaces and glossy objects with high specular regions. Additionally, we incorporate the shortest-axis normal modeling with depth-related regularization to represent reliable geometry, resulting in improved shape reconstruction. Furthermore, we enhance the probe-based baking solution to better model ambient occlusion and handle indirect illumination. Through extensive experiments on various challenging datasets, including complex scenes and glossy objects, we demonstrate the superiority of our method over state-of-the-art methods qualitatively and quantitatively. Our results highlight the efficacy of GUS-IR in achieving high-quality intrinsic decomposition, particularly for glossy surfaces.

\begin{figure}[t]
\centering
\captionsetup[subfloat]{labelfont=small,textfont=small}
\subfloat[Relighting of GUS-IR]{\includegraphics[width=0.45\linewidth]{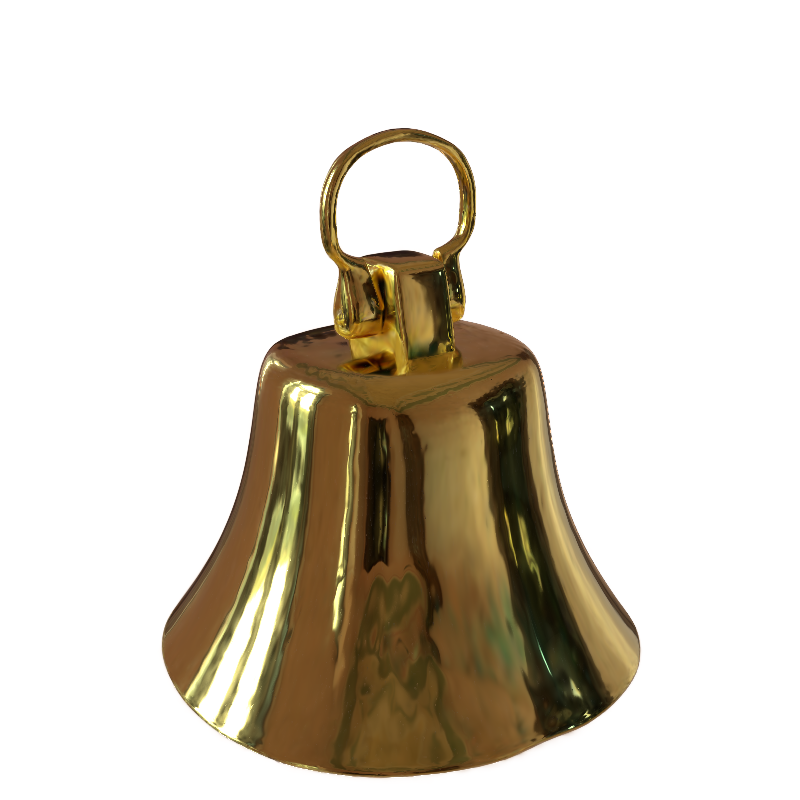}}
\subfloat[Relighting Reference]{\includegraphics[width=0.45\linewidth]{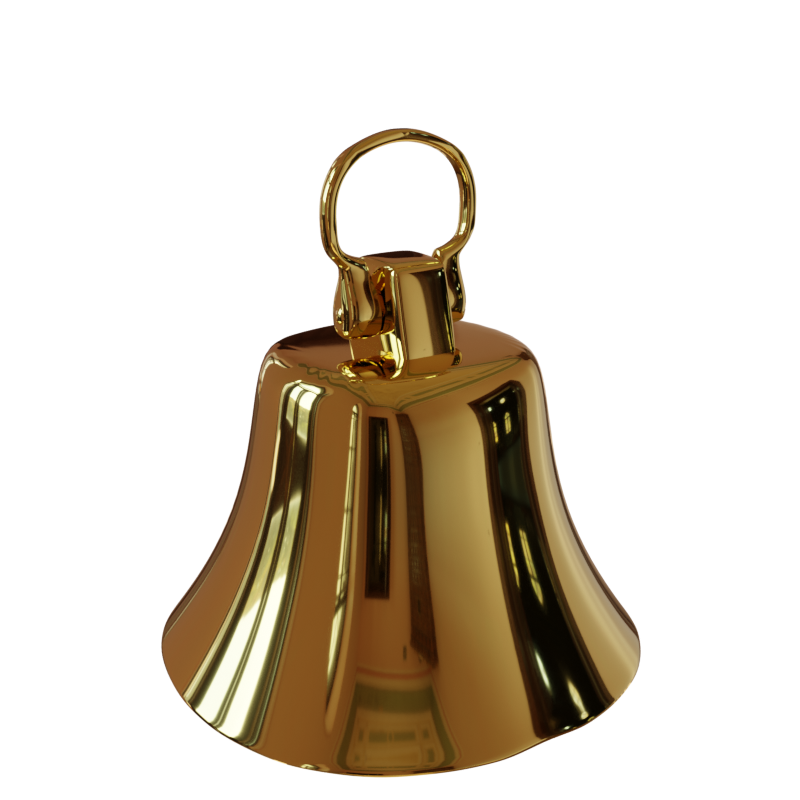}}
\caption{GUS-IR fails to estimate smooth and accurate normals on the waist body of the bell, leading to distorted relighting results.}
\label{fig:limitation}
\end{figure}

\noindent\textbf{Limitation}
In GUS-IR, we take the shortest axis as normal for each particle.
Although our scheme significantly outperforms the other two schemes (\textit{Attach Normal} and \textit{Residual Normal}), it still struggles to accurately reconstruct the normal of the glossy surfaces compared to the method based on implicit nerual field to construct SDF (\eg NeRO~\cite{liu2023nero}).
This leads to distorted relighting results on smooth and glossy surfaces as shown in \cref{fig:limitation}, damaging the relighting quality.
This is the reason that our method is notably inferior to NeRO on the Glossy Blender dataset in \cref{tab:relighting}.
We believe it is valuable to improve the geometry representation of 3DGS and address this limitation in future work.




\bibliographystyle{IEEEtran}
\bibliography{IEEEabrv,main}
%



\vspace{11pt}



\end{document}